\documentclass[11pt]{article}

\usepackage[margin=1in]{geometry}
\usepackage{amsmath,amssymb}
\usepackage{booktabs,array,microtype}
\usepackage{graphicx}
\usepackage{hyperref}
\usepackage{xcolor}
\usepackage{placeins}

\extrafloats{100}

\hypersetup{hidelinks}
\graphicspath{{figures/}}
\newcommand{\affilmark}[1]{\ensuremath{{}^{#1}}}

\title{Auditing Training-Free 3D Shape Retrieval with Diffused Geodesic Moments}
\author{
\begin{tabular}{c}
Zhicheng Du\affilmark{1},
Changyue Liu\affilmark{2},
Wenji Xi\affilmark{3},
Zhaotian Xie\affilmark{1},\\
Zhuo Deng\affilmark{1},
Ziheng Zhang\affilmark{1},
Yang Liu\affilmark{1},
Lan Ma\affilmark{1}\\[0.5em]
\small \affilmark{1}\,Tsinghua Shenzhen International Graduate School, Tsinghua University\\
\small \affilmark{2}\,Guangzhou International Economics College\\
\small \affilmark{3}\,School of Electrical and Electronic Engineering, The University of Sheffield
\end{tabular}
}
\date{May 2026}

\newcommand{\R}{\mathbb{R}}

\newcommand{\eps}{\varepsilon}
\newcommand{\best}[1]{\textbf{#1}}
\newcommand{\second}[1]{\underline{#1}}
\newcounter{algorithm}

\begin{document}
\maketitle

\begin{abstract}
Reported retrieval scores for training-free shape descriptors conflate local signal design, normalization, aggregation, codebook fitting, and metric choices, making isolated component evaluation difficult.  This paper reframes descriptor evaluation as a {\em protocol audit}.  We introduce Diffused Geodesic Moments (DGM), a seed-conditioned descriptor that computes sparse implicit heat responses, converts them to distance-like fields, and summarizes each vertex by low-order moments across seeds and scales.  DGM is used both as a practical non-spectral baseline and as an instrument for isolating protocol effects.  On the registered FAUST benchmark split (FAUST-Reg) and the TOSCA shape collection, aggregation-matched experiments show that an independent Geometric Moment Shape Descriptor baseline built on Heat Kernel Signature features (GMSD-HKS) obtains the highest scores in this implementation ($0.621/0.820$ and $0.865/0.963$ mean average precision (mAP)/top-1), Wave Kernel Signature (WKS) remains a strong classical signal, and DGM is useful mainly when sparse solves, non-spectral deployment, or symmetry-informative seed frames are priorities.  The broader finding is methodological: the input field and aggregation protocol can dominate the moment formula.  The paper contributes a reproducible protocol-cascade analysis, a cross-shape alignment diagnostic for functional-map compatibility, and concrete recommendations for designing and reporting training-free shape descriptors.
\end{abstract}

\section{Introduction}
Local descriptors for non-rigid shapes often begin with intrinsic operators on a mesh.  Classical signatures such as the Heat Kernel Signature (HKS), the Scale-Invariant Heat Kernel Signature (SI-HKS), and WKS rely on spectral responses of the Laplace--Beltrami operator \cite{sun2009hks,bronstein2010sihks,aubry2011wks}; functional map pipelines then employ such descriptors as probe functions or preservation constraints \cite{ovsjanikov2012fm}.  In retrieval, local descriptors are typically aggregated into global shape codes through bag-of-features, Shape Google, or Vector of Locally Aggregated Descriptors (VLAD)-style encodings \cite{bronstein2011shapegoogle,jegou2010vlad}.

Recent shape matching work has revisited spectral bias and frequency selection.  Frequency-aware functional maps learn spectral filters for difficult matching regimes \cite{luo2025dfafm}, while spectral-basis learning replaces fixed Laplacian eigenbases with learned ones \cite{luo2026spectralbasis}.  DGM takes a different route: instead of repairing a truncated spectral basis, it probes the surface directly in the spatial domain through seed-conditioned fields.  This avoids spectral truncation as an implementation bottleneck, while introducing a different one---which seeds, which scales, which field transforms, and which aggregations preserve useful geometric information?

The paper poses a specific question within this classical setting: if one forgoes a Laplacian eigendecomposition and instead computes a small bank of seed-conditioned heat responses, how much information do simple distributional moments of those fields carry?  The question is motivated by a gap in the literature.  HKS and WKS describe each point through diagonal spectral responses; geodesic distribution methods summarize distances to samples or across the shape.  DGM occupies the middle ground: each vertex is represented by statistics of its responses to multiple seed-conditioned, off-diagonal fields, at several diffusion scales.

The empirical answer is intentionally conservative.  DGM yields a compact and interpretable descriptor pipeline, and DGM-VLAD gives useful retrieval performance on several benchmarks.  Under aggregation-matched evaluation, however, WKS is stronger, and an independent GMSD-HKS implementation obtains the highest retrieval scores on FAUST-Reg and TOSCA in our implementation.  This outcome is the point of the audit: moment statistics are not decisive by themselves; the input signal and aggregation protocol are major algorithmic choices whose effects can reverse rankings.

The paper is therefore organized as a {\em protocol audit}.  A retrieval number is treated as the result of a cascade
\[
    \mathrm{Score}
    =
    \mathrm{Metric}\!\left(
    \mathrm{Aggregation}\!\left(
    \mathrm{Normalization}\!\left(
    \mathrm{LocalDescriptor}\!\left(
    \mathrm{Field}(S)\right)\right)\right)\right),
\]
rather than as a property of the local descriptor alone.  This framing makes negative results useful: when changing the aggregation or input field reverses method rankings, the descriptor literature ought to report those layers explicitly.

We also use this audit to make explicit the {\em descriptor--solver contract}: the assumptions a downstream solver makes about the descriptors it receives.  For functional maps, the contract requires at least two properties.  First, descriptor coordinates should be comparable across shapes, which we measure by the \textbf{C}ross-\textbf{S}hape \textbf{A}lignment \textbf{S}core (CSAS).  Second, useful descriptor energy should lie within the frequency band retained by the solver, which we measure by spectral compressibility.  A descriptor can be locally informative while still failing inside the solver if either part of this contract is violated.

\begin{figure}[t]
    \centering
    \includegraphics[width=\linewidth]{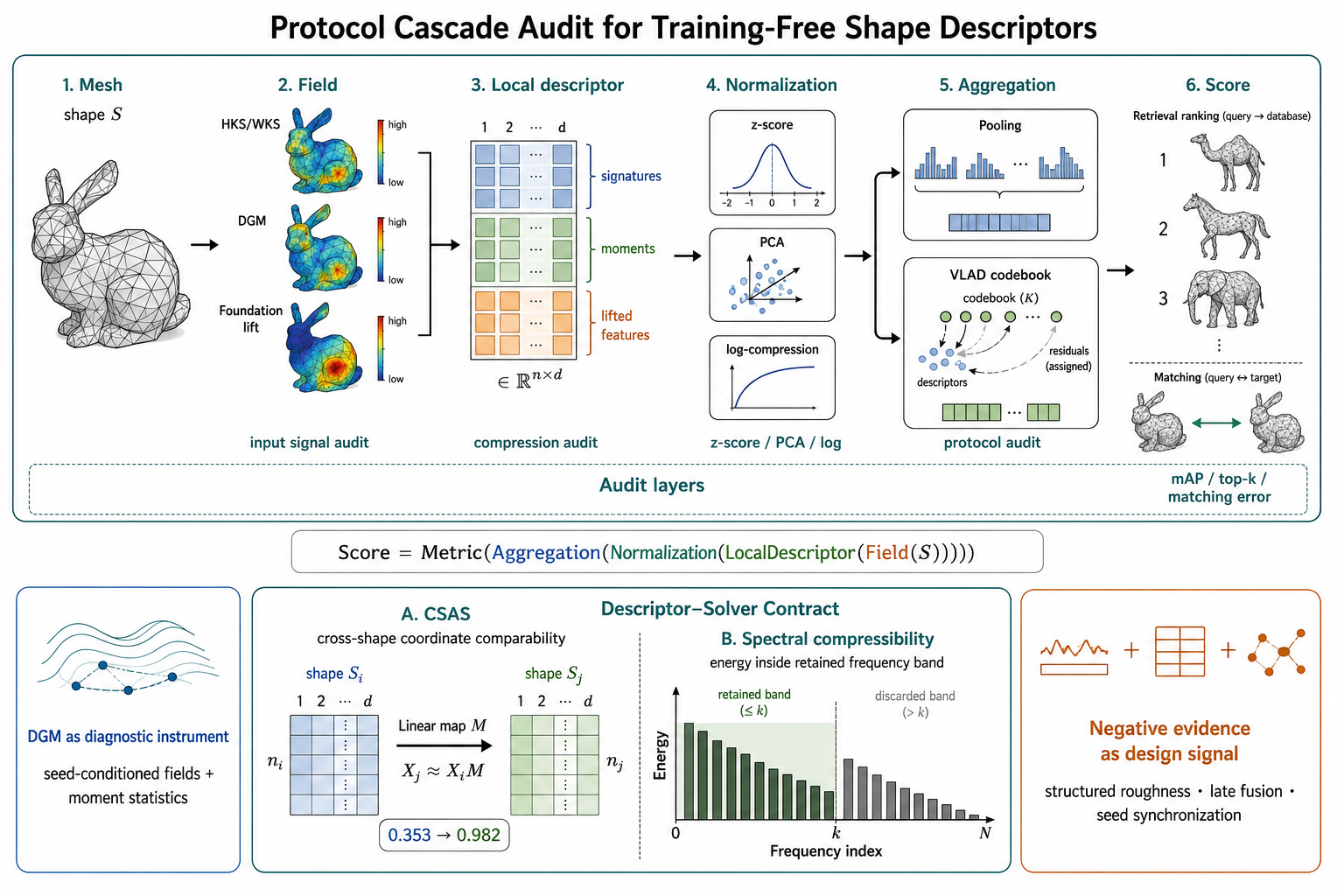}
    \caption{Protocol cascade audit.  The paper treats a retrieval or matching score as the output of a cascade of auditable choices: field construction, local descriptor design, normalization, aggregation, codebook fitting, metric, and solver interface.  DGM is the diagnostic instrument used to expose how seed-conditioned fields, moment compression, protocol choices, and descriptor--solver compatibility affect the final score.}
    \label{fig:protocol-audit}
\end{figure}

Figure~\ref{fig:protocol-audit} gives the visual organization of the paper.  DGM supplies the seed-conditioned field and moment-compression instrument, while the audit follows how later protocol choices change the reported score.

The contributions are organized around this audit.
First, we introduce a {\em protocol-cascade audit} for training-free shape retrieval, which decomposes a reported retrieval number into field construction, local statistics, normalization, pooling or VLAD aggregation, codebook fitting, dimensionality reduction, and the final metric.
Second, we develop {\em Diffused Geodesic Moments} as a seed-conditioned descriptor family that operates over off-diagonal heat and geodesic fields; it serves both as the audit instrument and as a practical non-spectral baseline.
Third, we provide a {\em reproducible comparison} against HKS, SI-HKS, WKS, GMSD-style spectral moments, non-eigendecomposition heat-kernel approximations, and controlled learned or lifted feature probes, all under fixed random seeds and documented protocols.
Fourth, we report {\em stability and interpretation diagnostics} for seed choice, vertex reordering, moment compression, soft Voronoi behavior, symmetry-informative side information, cross-shape descriptor alignment, and functional-map compatibility.  These diagnostics test the descriptor--solver contract: a descriptor may be locally discriminative while still violating the coordinate-comparability and spectral-bandwidth assumptions of a downstream solver.
Finally, the audit yields {\em concrete design recommendations}: report both native and aggregation-matched scores, repeat fitted codebooks, treat the input field and aggregation layer as first-class algorithmic choices, and test descriptor--solver compatibility before claiming matching performance.

\section{Related Work and Positioning}
\paragraph{Spectral point signatures.}
HKS restricts the heat kernel to the diagonal $h_t(x,x)$ and samples it across diffusion scales \cite{sun2009hks}.  SI-HKS builds scale invariance by transforming the logarithmic scale axis \cite{bronstein2010sihks}.  WKS uses band-pass filters over Laplacian eigenfunctions and is often strong in retrieval and matching \cite{aubry2011wks}.  Recent functional-map methods further expose the importance of frequency selection through learned spectral filters and learned bases \cite{luo2025dfafm,luo2026spectralbasis}.  In the baseline implementation used here, classical spectral descriptors are computed from a partial eigendecomposition; this is a common but not mandatory implementation choice.

\paragraph{Heat kernel approximations and geodesics.}
Eigendecomposition is not the only way to evaluate heat-related quantities.  Efficient alternatives include multi-resolution heat-kernel approximation on surfaces \cite{vaxman2010heatkernels}, rational or polynomial approximations of the heat semigroup, and direct heat-flow distance methods such as Geodesics in Heat \cite{crane2013geodesics}.  The default DGM field is closest in spirit to a short-time heat-flow construction, but it does not compute the two-Poisson-solve heat method distance.  It uses a regularized implicit response followed by a normalized log transform.  We include Chebyshev-Hutchinson, Pad\'e/backward-Euler-Hutchinson, coarse multiresolution graph-proxy HKS baselines, and a heat-method geodesic field diagnostic as controlled comparisons.  These controlled implementations are included to separate speed from retrieval accuracy in this pipeline; they should not be read as optimized reference implementations of the full heat-kernel approximation literature.

\paragraph{Moment descriptors.}
The recent Geometric Moment Shape Descriptor (GMSD) work by Zhang, Liu, Wu, Lv, and Zhao \cite{zhang2026gmsd} is the closest modern reference for moment-based shape descriptors.  GMSD computes geometric moments of spectral shape descriptors and reports strong robustness and correspondence/retrieval results.  The distinction is the input signal: DGM moments are taken over seed-conditioned heat/geodesic fields, while GMSD moments are taken over spectral descriptor responses.  We implement a six-term GMSD-style baseline from the public definition: temporal mean, variance, and skewness over the signature scale axis, plus one-ring conditional expectation and variance terms.  This independent implementation is not the authors' released code, but it provides a direct test of whether spectral moment descriptors explain the retrieval behavior.

Luciano and Ben Hamza \cite{luciano2018deepgm} are an important earlier reference for geodesic moment descriptors.  Their DeepGM framework uses geodesic moments for 3D shape classification and then learns high-level features with stacked sparse autoencoders.  That work already connects geodesic moments, heat-kernel intuition, and spectral geometry.  DGM follows the same broad idea that moments of distance-like functions are informative, but differs in three concrete ways: the input is a bank of seed-conditioned off-diagonal heat/geodesic fields, the descriptor is local before pooling/VLAD aggregation, and the experiments isolate the effect of moments from the field and aggregation protocol rather than training a classifier on top of the moments.

\paragraph{Geodesic statistics and shape retrieval.}
Geodesic distributions have a long history.  Rabin, Peyre, and Cohen use optimal mass transport for geodesic shape retrieval \cite{rabin2010omt}; Ion et al. use geodesic eccentricity for 3D shape matching \cite{ion2008eccentricity}.  These works make clear that statistics over distance fields are not new.  DGM sits in this family but uses per-vertex moments across multiple seed-conditioned fields and then optionally aggregates local descriptors through VLAD.

\paragraph{Aggregation and global retrieval.}
VLAD aggregates local descriptor residuals against a codebook \cite{jegou2010vlad}.  In 3D shape retrieval, Shape Google and related bag-of-features pipelines showed how diffusion descriptors can be converted into global retrieval codes \cite{bronstein2011shapegoogle}.  This motivates one of the central protocol choices of this paper: a descriptor should be tested both in its native pipeline and under a matched aggregation family.

\paragraph{Learning and foundation features.}
Learning-based surface descriptors and recent foundation-feature pipelines now form a separate and often stronger regime.  DiffusionNet learns diffusion-based surface features that are robust to discretization changes \cite{sharp2022diffusionnet}.  Diff3F lifts DINO (self-distillation with no labels) and diffusion-model features from rendered views to untextured 3D surfaces and reports zero-shot semantic correspondence results \cite{dutt2024diff3f}.  DiffuMatch learns category-agnostic spectral diffusion priors for robust non-rigid matching \cite{pierson2025diffumatch}.  Synchronous Diffusion regularizes unsupervised matching by enforcing diffusion consistency across shapes \cite{cao2024synchronous}.  LiteGE constructs lightweight geodesic embeddings for efficient geodesic computation and non-isometric correspondence \cite{adikusuma2026litege}.  More broadly, recent vision foundation models such as DINOv3 emphasize high-quality dense image features \cite{simeoni2025dinov3}, and a 2026 state-of-the-art survey identifies foundation features and partial matching as active opportunities in non-rigid correspondence \cite{zhuravlev2026star}.  DGM is a classical, training-free counterpart to this learned family: it uses hand-designed diffusion fields as probe signals and evaluates how far such signals go without learned priors.

\paragraph{Information and topology viewpoints.}
Information-bottleneck interpretations of representation learning emphasize the tradeoff between preserved task information and representation complexity \cite{katende2025vgib}.  DGM is not a learned bottleneck, but its design creates a measurable compression step: a $k$-seed response distribution is reduced to six moment channels per scale.  Persistent-homology work on 3D shapes provides another complementary view, using multiscale topological summaries rather than only metric responses \cite{kudeshia2026topogat,kudeshia2026phdesign}.  We use these literatures as diagnostic motivation, not as claims that DGM optimizes an information-bottleneck objective or computes topological invariants.

\paragraph{Hybrid intrinsic/extrinsic descriptors.}
Recent descriptors also combine intrinsic and extrinsic cues, for example AWEDD, a descriptor that jointly encodes multiscale extrinsic and intrinsic shape features \cite{liu2023awedd}.  This is relevant because purely intrinsic descriptors can be ambiguous under symmetries.  Seed-free spectral signatures such as HKS and WKS are naturally symmetry-agnostic, whereas seed-conditioned descriptors introduce a frame that can become symmetry-informative; recent symmetry-disentanglement work treats this distinction as an explicit representation choice \cite{weissberg2026symmetry}.  DGM is not fully intrinsic when Euclidean farthest-point sampling (FPS) or tensor invariants are enabled, and this is not only a limitation; it is also part of why some retrieval settings may improve.  We therefore report the distinction explicitly rather than describing the method as fully intrinsic.

Table~\ref{tab:positioning} summarizes the positioning used throughout the paper.  Its purpose is not to rank descriptor families, but to separate the source signal, the use of eigendecomposition, the presence of a seed frame, and the role each family plays in the audit.

\begin{table}[!htbp]
\centering
\footnotesize
\caption{Positioning of descriptor families discussed in this paper. The table separates the input signal from the statistics and aggregation stages, which is the main distinction needed to compare DGM with GMSD and classical spectral signatures.}
\label{tab:positioning}
\begin{tabular}{@{}>{\raggedright\arraybackslash}p{0.22\linewidth}>{\raggedright\arraybackslash}p{0.31\linewidth}cc>{\raggedright\arraybackslash}p{0.21\linewidth}@{}}
\toprule
family & primary signal & eig.? & seed? & role in this paper \\
\midrule
HKS / SI-HKS / WKS & pointwise spectral response & yes & no & canonical spectral baselines \\
GMSD / moment spectral descriptors & moments of spectral signatures & yes & no & closest moment prior art \\
Geodesic distribution / eccentricity descriptors & geodesic maps/statistics & no & often & geodesic-statistics prior art \\
DGM (this work) & seed-conditioned heat/\allowbreak geodesic fields & no & yes & seed-field descriptor study \\
\bottomrule
\end{tabular}
\end{table}

\section{Method}
\subsection{Notation}
Let a triangle mesh be $S=(V,F)$ with $n=|V|$ vertices in $\R^3$.  We build the cotangent stiffness matrix $L$ and lumped mass matrix $M$.  A DGM descriptor uses $k$ seed vertices $\mathcal{S}=\{s_1,\ldots,s_k\}$ and a set of diffusion scales $\mathcal{T}$.  Unless otherwise stated, meshes are area-normalized by the benchmark preprocessing code.

\subsection{Seed Selection}
A seed-conditioned descriptor necessarily depends on how the seed set is selected.  We therefore separate three regimes.  Random Euclidean farthest-point sampling is used only as a variance probe.  Deterministic Euclidean sampling chooses the first seed as the vertex farthest from the Euclidean centroid and resolves ties by coordinate-lexicographic order.  Deterministic graph-geodesic sampling performs farthest-point sampling on the mesh edge graph, initialized by a double-sweep diameter endpoint heuristic with the same deterministic tie breaking.
The graph-geodesic mode is a stronger default for an intrinsic-oriented descriptor, but it is still not mathematically canonical on exactly symmetric shapes: if several vertices are intrinsically indistinguishable, any seed-based descriptor must either break the symmetry or explicitly quotient it out.

\subsection{Seed-Conditioned Fields}
For a seed $s$ and scale $t$, DGM solves the regularized implicit system
\begin{equation}
    (M + tL + \eps I) u_{s,t} = e_s ,
    \label{eq:heat-solve}
\end{equation}
where $e_s$ is the canonical vector at vertex $s$ and $\eps=10^{-8}$ is a numerical diagonal regularizer used for stable sparse factorization.  This equation defines the practical implicit heat-response field used by DGM.  It is not identical to every finite-element convention for a Dirac heat source, and it is not a full heat-kernel approximation method.

Let $q_{95}(a)$ be the empirical $95$th percentile of a vector $a$, and let $[r]_+=\max(r,0)$ denote the positive part.  The response is converted to a distance-like field by
\begin{equation}
    \phi_{s,t}(x) =
    \frac{\psi_{s,t}(x)-\min_{y\in V} \psi_{s,t}(y)}
    {q_{95}(\psi_{s,t}-\min_{y\in V}\psi_{s,t})+10^{-12}},
    \qquad
    \psi_{s,t}(x)= -\log\left(
    \frac{[u_{s,t}(x)]_+}{\max_{y\in V} u_{s,t}(y)+10^{-12}} + 10^{-12}
    \right).
    \label{eq:proxy}
\end{equation}
This is a Varadhan-style monotone transform at finite time.  The positive part in Eq.~\ref{eq:proxy} is needed because cotangent discretizations are not, in general, positivity-preserving M-matrices under the point-source right-hand side used here.  The percentile normalization fixes cross-shape dynamic range while keeping large but valid field values from dominating all moment channels.

We also provide graph-geodesic and heat-method geodesic field modes where seed distances are transformed by $\log(1+d/t)$ and normalized.  The heat-method variant uses an external implementation of the heat-method distance solver.  These modes are included to separate the contribution of field choice from the moment and aggregation choices.

\begin{figure}[t]
    \centering
    \includegraphics[width=\linewidth]{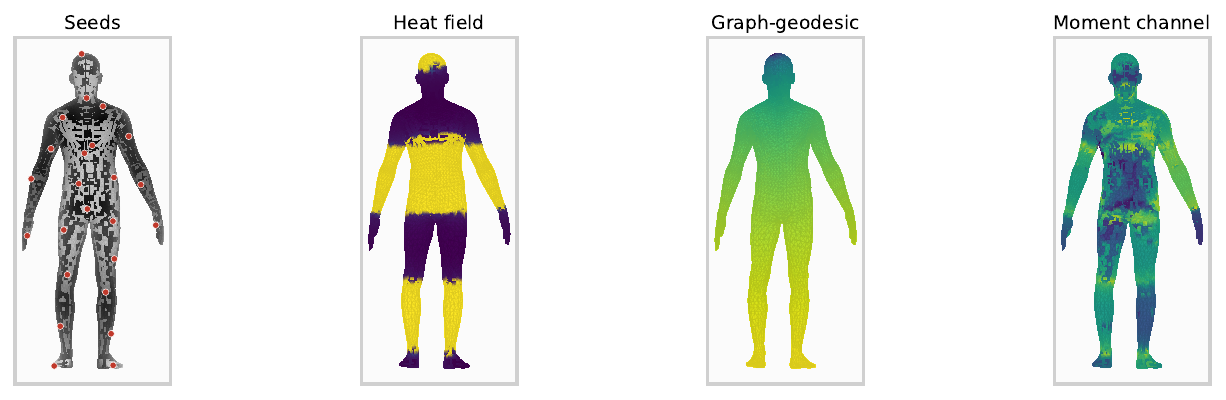}
    \caption{Representative DGM fields and channels.  The visualization shows that the descriptor is built from moments over seed-conditioned fields rather than from a diagonal spectral signature.}
    \label{fig:field-channels}
\end{figure}

Figure~\ref{fig:field-channels} illustrates the fields and moment channels used by the descriptor.  The figure is meant to clarify that DGM is built from seed-conditioned off-diagonal responses rather than from only diagonal heat signatures.

\subsection{Moments and Local Descriptors}
For a fixed vertex $x$ and scale $t$, collect the seed-conditioned field values
\[
    z_t(x) = \{\phi_{s,t}(x): s \in \mathcal{S}\}.
\]
DGM stores low-order statistics of this set:
\[
    m_t(x) = [\mu, \sigma^2, \operatorname{skew}, \operatorname{kurtosis},
    \min, \max](z_t(x)).
\]
The channels have direct geometric interpretations.  The mean approximates average distance-like response to the seed set.  The variance measures how unevenly a point sees the seed set and often separates central regions from extremities.  Skewness and kurtosis capture whether the response is dominated by a few nearby or far seeds.  The minimum is a nearest-seed response, while the maximum is related to eccentricity with respect to the chosen seed set.  The local descriptor concatenates $m_t(x)$ over all $t\in\mathcal{T}$, followed by feature-wise z-score normalization and signed-log compression to reduce heavy-tailed ranges after the log transform.

A useful geometric interpretation is a soft Voronoi diagram over the seed set.  For a temperature $\tau>0$, define
\[
    p_{s,t,\tau}(x)
    =
    \frac{\exp(-\phi_{s,t}(x)/\tau)}
    {\sum_{s'\in\mathcal{S}}\exp(-\phi_{s',t}(x)/\tau)} .
\]
At small scale or small temperature, most mass is assigned to the nearest seed and the field resembles a hard Voronoi cell assignment.  At larger diffusion scales, several seeds can have comparable responses and vertices near cell interfaces receive higher entropy assignments.  The DGM moments summarize the same seed-response distribution from a metric side: mean and variance describe centrality and boundary behavior, while min/max and higher moments describe nearest/farthest seed contrast and concentration.

Algorithm~\ref{alg:dgm} summarizes the extraction pipeline, including seed selection, sparse solves, moment construction, normalization, and optional global aggregation.

\begin{table}[t]
\refstepcounter{algorithm}
\centering
\small
\label{alg:dgm}
\begin{tabular}{p{0.16\linewidth}p{0.76\linewidth}}
\toprule
\multicolumn{2}{l}{\textbf{Algorithm \thealgorithm: Diffused Geodesic Moments}} \\
\midrule
\textbf{Input} & Mesh $(V,F)$, scale set $\mathcal{T}$, seed count $k$, moment channels $\mathcal{M}$ \\
\textbf{Output} & Vertex descriptors $D_{\mathrm{local}}$ and an optional global descriptor $D_{\mathrm{global}}$ \\
\midrule
1 & Assemble the cotangent stiffness matrix $L$ and lumped mass matrix $M$. \\
2 & Select seeds $\mathcal{S}=\{s_i\}_{i=1}^k$ by deterministic Euclidean or graph-geodesic farthest-point sampling. \\
3 & For each scale $t\in\mathcal{T}$, factor the sparse matrix $M+tL+\eps I$ once. \\
4 & For each seed $s\in\mathcal{S}$, solve $(M+tL+\eps I)u_{s,t}=e_s$ and convert the response to $\phi_{s,t}$ by Eq.~\ref{eq:proxy}. \\
5 & For each vertex $x$, compute moments of $\{\phi_{s,t}(x):s\in\mathcal{S}\}$ at every scale. \\
6 & Concatenate the moment channels over scales and apply the fixed descriptor normalization. \\
7 & Aggregate local descriptors by pooling or by VLAD when a global retrieval code is required. \\
\midrule
\multicolumn{2}{p{0.92\linewidth}}{\emph{Protocol note.} The final aggregation is included because the audit evaluates pooling and VLAD as distinct algorithmic choices rather than as neutral reporting details.} \\
\bottomrule
\end{tabular}
\end{table}

\subsection{Global Descriptors}
We evaluate two global encodings.  The first pools local descriptors by mean, standard deviation, and maximum.  The second uses VLAD: local descriptors are assigned to a learned codebook and residuals are concatenated, normalized, and used as the global retrieval vector.  The codebook protocol is frozen within each evaluation split so that DGM, HKS, WKS, and SI-HKS can be compared under the same aggregation family.

An optional tensor branch computes coordinate-weighted covariance invariants from the seed fields.  This branch is extrinsic and is disabled in some ablations.  Since it uses embedded coordinates, any result with the branch enabled must be described as intrinsic-oriented rather than fully intrinsic.

\paragraph{Implementation defaults.}
The main FAUST-Reg and TOSCA runs use 24 seeds, four DGM scales $\{0.01,0.03,0.07,0.15\}$, a six-channel moment set, 48 spectral eigenpairs for HKS/WKS/SI-HKS baselines, 24 HKS times or WKS bins, VLAD with 16 clusters, cosine retrieval, and principal component analysis (PCA) target dimension 96 in the fair runs.  The heat-approximation baselines use the same four reported scales, 24 target times, 8 Hutchinson probes, Chebyshev degree 24, 4 backward-Euler/Pad\'e steps, and 384 landmarks for the multiresolution graph proxy.  The heat-method geodesic field diagnostic uses the same seed count, moments, VLAD setting, and projection target.  The runtime random seed is fixed to 13 unless an experiment explicitly varies it.

\subsection{Complexity}
For $|\mathcal{T}|$ scales and $k$ seeds, DGM performs one sparse factorization of $M+tL+\eps I$ per scale and then $k$ triangular solves per scale.  If the mesh has $n$ vertices and factorization cost is $F(n)$ with memory fill-in dependent on mesh ordering, the dominant cost is
\[
    \sum_{t\in\mathcal{T}} F_t(n) + |\mathcal{T}|\, k\, S_t(n),
\]
where $S_t(n)$ is the cost of one sparse back-substitution.  With $r$ retained eigenpairs, partial Arnoldi/Lanczos methods require sparse matrix-vector products plus orthogonalization, with a typical orthogonalization term of order $O(r^2 n)$ in addition to operator applications and factorization/preconditioning costs.  Table~\ref{tab:efficiency} compares descriptor extraction times against the eigendecomposition-based baselines in this repository; optimized multi-resolution, rational, polynomial, prefactored, and graphics processing unit (GPU) heat-kernel implementations define different engineering points.

\begin{table}[!htbp]
\centering
\small
\caption{Average descriptor runtime per shape. Runtime is reported in seconds per shape and is lower-is-better ($\downarrow$). The spectral baselines are the eigendecomposition-based implementations in this repository.}
\label{tab:efficiency}
\begin{tabular}{lcccc}
\toprule
dataset & DGM sec./shape $\downarrow$ & HKS sec./shape $\downarrow$ & WKS sec./shape $\downarrow$ & SI-HKS sec./shape $\downarrow$ \\
\midrule
faust\_reg & \best{0.655} & 2.059 & 1.972 & \second{1.925} \\
tosca & \best{3.377} & \second{7.416} & 7.578 & 7.649 \\
kids & \best{4.665} & \second{7.252} & 7.485 & 7.269 \\
shrec20b & \best{2.835} & \second{4.223} & 4.236 & 4.502 \\
\bottomrule
\end{tabular}
\end{table}

\section{Properties and Interpretation}
\paragraph{Proposition 1: isometry behavior.}
Consider two triangulated surfaces related by an exact intrinsic isometry that preserves the cotangent Laplacian, mass matrix, scale set, and seed set up to permutation.  The linear systems in Eq.~\ref{eq:heat-solve} commute with the induced vertex permutation.  Therefore the seed-conditioned fields are permuted in the same way, and permutation-invariant statistics over the seed axis give identical DGM moment channels at corresponding vertices.  This gives DGM the expected isometry behavior when seed selection is itself intrinsic and deterministic.

\paragraph{Proposition 2: seed sampling limit.}
For a fixed vertex $x$ and scale $t$, the moment channels are empirical moments of the random variable $\Phi_t(x,S)=\phi_{S,t}(x)$, where $S$ is a seed sampled from a seed distribution on the surface.  If the seeds are dense samples from a distribution $\rho$, the empirical mean, variance, skewness, and kurtosis converge to the corresponding moments of $\Phi_t(x,S)$ under $\rho$ by the law of large numbers.  In the continuous limit, DGM describes each point by the distribution of distance-like responses from the whole seed measure, rather than from a finite seed set.  A finite deterministic FPS set is a quadrature rule for this distribution; it improves coverage but necessarily fixes a coordinate frame on symmetric shapes.

\paragraph{Proposition 3: symmetry breaking.}
Seed-conditioned descriptors cannot be both seed-dependent and invariant to all intrinsic automorphisms.  If a shape has a nontrivial symmetry group and the selected seed set is not invariant under a symmetry, two intrinsically equivalent vertices can receive different descriptors.  Deterministic seed rules remove vertex-index dependence but still choose one representative among symmetric alternatives.  The seed-stability experiments quantify this tradeoff rather than treating it as a hidden implementation detail.

\paragraph{Finite-time heat transform.}
For the continuous heat kernel $h_t(x,y)$, Varadhan's formula states that $-4t\log h_t(x,y)$ approaches squared geodesic distance as $t\to0$.  DGM uses a normalized finite-time resolvent response rather than an exact heat kernel, so Eq.~\ref{eq:proxy} is best read as a monotone distance-like field.  As $t\to0$, the response concentrates around the seed and the log field separates near and far neighborhoods.  As $t$ grows, the response becomes flatter and moment channels collapse.  The scale set therefore controls a bias-variance tradeoff: small scales emphasize local structure and discretization sensitivity, while larger scales emphasize coarse placement and eventually lose contrast.

\paragraph{Relation to spectral descriptors.}
In the continuous setting, heat responses admit the expansion $h_t(x,s)=\sum_i e^{-t\lambda_i}\varphi_i(x)\varphi_i(s)$.  HKS keeps only the diagonal $h_t(x,x)$, while DGM summarizes the off-diagonal slice $s\mapsto h_t(x,s)$ through moments over seeds.  If seeds are dense and uniformly weighted, the first raw moment of the unnormalized heat response integrates the heat kernel over the surface and approaches a constant on a closed connected manifold.  This explains why DGM relies on nonlinear log/normalization and higher moments: the discriminative information is not in the total heat mass but in the distribution of seed-conditioned responses.  GMSD-HKS instead applies moments to spectral point signatures, which the experiments show can be the stronger signal under VLAD.

\paragraph{Information-compression view.}
With $k=24$ seeds, each scale first produces a 24-dimensional response distribution at every vertex and then compresses it to six moment channels.  This is a fixed geometric bottleneck rather than a learned encoder.  The information-bottleneck analogy is useful only if the retained information is measured, so we report PCA and regression diagnostics instead of asserting minimal sufficiency.  The diagnostic asks whether the six moment channels preserve the sorted seed-response distribution well enough to explain the retrieval behavior, and it records where the compression is visibly lossy.

\section{Experimental Protocol}
The audit is structured around controlled comparisons that isolate one layer of the protocol cascade at a time.  We first describe the datasets, then define the native and fair evaluation protocols, and finally introduce the diagnostic probes that test specific properties of the descriptor pipeline.

\paragraph{Metric notation.}
Tables use arrows only next to metric names: $\uparrow$ means larger is better and $\downarrow$ means smaller is better.  When a table has a clear ranking within a dataset, row, or protocol block, bold numbers mark the best value and underlined numbers mark the second-best value; ties receive the same mark.  Retrieval quality is reported by mAP and top-1 retrieval accuracy.  Matching diagnostics report geodesic error, hit@10\% diameter, nearest-neighbor (NN) error, and CSAS.  Runtime is reported in seconds per shape, and robustness drops are mAP decreases from the clean run, so smaller drops are better.  Compact table headers use common abbreviations only for space: err. for error, acc. for accuracy, rel. for relative, dim. for dimension, sec./shape for seconds per shape, geo. for geodesic, spec. for spectral, and GT for ground truth.

\paragraph{Datasets.}
Retrieval experiments use processed FAUST-Reg, TOSCA, Kids, and SHREC20B manifests.  FAUST-Reg denotes the registered split of the FAUST benchmark and uses the 100 registered training meshes \cite{bogo2014faust}.  TOSCA is the high-resolution non-rigid shape collection commonly used for shape similarity and correspondence experiments \cite{bronstein2008numerical}.  Kids is the deformable shape matching dataset from Rodola et al. \cite{rodola2014dense}.  SHREC20B denotes subset B of the Shape Retrieval Contest 2020 (SHREC'20) non-isometric correspondence benchmark and uses its high-resolution models \cite{dyke2020shrec20}.  FAUST-Reg and TOSCA are the main non-rigid retrieval benchmarks.  Kids and SHREC20B are included as native implementation checks.  Table~\ref{tab:datasets} lists the role of each processed dataset in the audit.  SCAPE is not used for class retrieval because the processed registered meshes form a single category; its manifest is prepared for release and correspondence-oriented follow-up experiments only.

\begin{table}[!htbp]
\centering
\small
\caption{Datasets and protocols used in the experiments.}
\label{tab:datasets}
\begin{tabular}{@{}p{0.12\linewidth}p{0.30\linewidth}cp{0.12\linewidth}p{0.26\linewidth}@{}}
\toprule
dataset & task & shapes & ids/classes & protocol \\
\midrule
faust\_reg & non-rigid retrieval; registered check & 100 & 10 & native, fair, correspondence diagnostic \\
tosca & non-rigid retrieval; robustness & 80 & 9 & native, fair, robustness \\
kids & fine-grained retrieval & 32 & 2 & native retrieval \\
shrec20b & heterogeneous retrieval & 14 & 11 & native retrieval \\
scape & registered extraction/scaling check & 71 & 1 & no class retrieval; one-label manifest \\
\bottomrule
\end{tabular}
\end{table}

\paragraph{Native implementation protocol.}
Each descriptor is evaluated with its default implementation path in this codebase.  This protocol is useful for end-to-end engineering comparison, but it mixes descriptor differences and aggregation choices.  Table~\ref{tab:native} reports this native protocol before any aggregation-matched comparison is imposed.

\begin{table}[!htbp]
\centering
\small
\caption{Native implementation retrieval mAP ($\uparrow$). These numbers use each descriptor in the implementation provided by the benchmark code.}
\label{tab:native}
\begin{tabular}{lcccc}
\toprule
method & FAUST-Reg mAP $\uparrow$ & TOSCA mAP $\uparrow$ & Kids mAP $\uparrow$ & SHREC20B mAP $\uparrow$ \\
\midrule
DGM-VLAD & \best{0.344} & 0.575 & \best{0.891} & 0.159 \\
DGM & 0.230 & 0.388 & 0.668 & 0.095 \\
HKS & 0.261 & \second{0.725} & \second{0.851} & \best{0.176} \\
WKS & \second{0.305} & \best{0.763} & 0.792 & \second{0.163} \\
SI-HKS & 0.196 & 0.466 & 0.570 & 0.158 \\
\bottomrule
\end{tabular}
\end{table}

\paragraph{Fair aggregation protocol.}
For FAUST-Reg and TOSCA, all descriptors are rebuilt with the same pooling or VLAD aggregation family and evaluated with the same metric.  This is the main protocol for judging the descriptor signal itself.  Table~\ref{tab:fair-vlad} shows why this distinction matters: DGM-VLAD improves over HKS-VLAD on FAUST-Reg in the four-method matched run, but WKS-VLAD is much stronger than DGM on both FAUST-Reg and TOSCA.

\begin{table}[!htbp]
\centering
\small
\caption{Aggregation-matched VLAD retrieval.  Both mAP and top-1 are higher-is-better ($\uparrow$).  This is the main fair comparison because all descriptors share the same aggregation family and evaluation metric.}
\label{tab:fair-vlad}
\begin{tabular}{lcc}
\toprule
method & FAUST-Reg mAP$\uparrow$/top-1$\uparrow$ & TOSCA mAP$\uparrow$/top-1$\uparrow$ \\
\midrule
DGM & \second{0.360} / \second{0.530} & 0.533 / 0.750 \\
HKS & 0.297 / 0.410 & 0.608 / 0.812 \\
WKS & \best{0.602} / \best{0.850} & \best{0.835} / \second{0.950} \\
SI-HKS & 0.335 / 0.400 & \second{0.784} / \best{0.963} \\
\bottomrule
\end{tabular}
\end{table}

\paragraph{Moment and heat-approximation baselines.}
Table~\ref{tab:extended-baselines} includes baselines that separate three sources of performance: spectral input signals, moment statistics, and heat-kernel computation.  GMSD-HKS and GMSD-WKS test whether moment statistics over standard spectral signatures already explain the retrieval gains.  HKS-Cheb estimates the heat-kernel diagonal with Chebyshev filtering and Hutchinson probes; HKS-Pad\'e uses repeated backward-Euler, i.e. Pad\'e $[0/1]$, heat steps with Hutchinson probing; HKS-MR proxy builds a landmark graph and interpolates a coarse graph heat signature.  These three heat-kernel approximation baselines avoid a full spectral basis, but they are controlled implementations rather than optimized replacements for the cited reference methods.  Appendix Table~\ref{tab:extended-timing} reports their extraction time, while Table~\ref{tab:efficiency} reports the native extraction time of the main descriptor families.  The approximation baselines are cheaper, but Table~\ref{tab:extended-baselines} shows that this efficiency does not translate to competitive retrieval in the tested implementation.  The DGM field ablation also includes a heat-method geodesic input field, so that the descriptor is compared against a direct heat-flow distance computation rather than only against graph shortest paths.

\begin{table}[!htbp]
\centering
\small
\caption{Extended VLAD-cosine baselines for separating moment statistics, spectral input signals, and heat-kernel approximations.  Both mAP and top-1 are higher-is-better ($\uparrow$).  GMSD rows are independent six-moment implementations over HKS/WKS signatures; HKS-Pad\'e, HKS-Cheb, and HKS-MR proxy avoid a full spectral basis but are controlled implementations rather than optimized reference solvers.}
\label{tab:extended-baselines}
\begin{tabular}{lcc}
\toprule
method & FAUST-Reg mAP$\uparrow$/top-1$\uparrow$ & TOSCA mAP$\uparrow$/top-1$\uparrow$ \\
\midrule
GMSD-HKS & \best{0.621} / \second{0.820} & \best{0.865} / \second{0.963} \\
WKS & \second{0.602} / \best{0.850} & \second{0.835} / 0.950 \\
GMSD-WKS & 0.465 / 0.690 & 0.799 / \best{0.975} \\
SI-HKS & 0.335 / 0.400 & 0.784 / \second{0.963} \\
DGM & 0.360 / 0.530 & 0.533 / 0.750 \\
HKS & 0.297 / 0.410 & 0.608 / 0.812 \\
HKS-Pad\'e & 0.217 / 0.300 & 0.446 / 0.688 \\
HKS-MR proxy & 0.205 / 0.240 & 0.432 / 0.637 \\
HKS-Cheb & 0.152 / 0.130 & 0.264 / 0.175 \\
\bottomrule
\end{tabular}
\end{table}

\paragraph{Robustness protocol.}
Robustness is evaluated by reporting severity-wise retrieval drops under a frozen codebook.  Figure~\ref{fig:robustness-curves} summarizes the drop curves for partiality and decimation, while Table~\ref{tab:robustness-fixed} reports one fixed severity per perturbation for partiality, decimation, remeshing, and noise.  A negative drop means the perturbed run scored slightly higher; this happens for WKS-VLAD on remeshing and for DGM-VLAD on TOSCA noise, so the correct interpretation is robustness nuance rather than a universal win.

\begin{table}[!htbp]
\centering
\small
\caption{Fixed-severity robustness, reported as mAP drop from the clean run ($\downarrow$). Negative values mean the perturbed run scored slightly higher.}
\label{tab:robustness-fixed}
\begin{tabular}{llccc}
\toprule
dataset & perturb. & DGM-VLAD drop $\downarrow$ & HKS-VLAD drop $\downarrow$ & WKS-VLAD drop $\downarrow$ \\
\midrule
faust\_reg & partial & \second{0.234} & \best{0.166} & 0.470 \\
faust\_reg & decimation & \second{0.231} & \best{0.158} & 0.480 \\
faust\_reg & remeshing & \second{0.014} & 0.051 & \best{-0.035} \\
faust\_reg & noise & \second{0.244} & \best{0.171} & 0.470 \\
tosca & partial & \best{0.316} & \second{0.374} & 0.598 \\
tosca & decimation & \second{0.332} & \best{0.258} & 0.621 \\
tosca & remeshing & \second{0.040} & 0.052 & \best{-0.009} \\
tosca & noise & \best{-0.019} & \second{0.083} & 0.151 \\
\bottomrule
\end{tabular}
\end{table}

\paragraph{Seed stability protocol.}
We run seed-count and seed-initialization sweeps on FAUST-Reg with $k\in\{8,16,24,32\}$, five random starts for the Euclidean random mode, and one deterministic run for each deterministic mode.  We also run a confirmatory TOSCA sweep at $k\in\{16,24\}$ with three random starts and the same deterministic alternatives.  Vertex reordering is tested by recomputing descriptors after a random permutation and aligning local descriptors back to the original order.  Table~\ref{tab:seed-stability} reports the main sweep.

\begin{table}[!htbp]
\centering
\small
\caption{Seed-count and seed-initialization stability for DGM-VLAD.  mAP is higher-is-better ($\uparrow$).}
\label{tab:seed-stability}
\begin{tabular}{lllcc}
\toprule
dataset & seed mode & k & runs & mAP $\uparrow$ \\
\midrule
faust\_reg & euclidean\_deterministic & 8 & 1 & 0.240 \\
faust\_reg & euclidean\_random & 8 & 5 & 0.253 $\pm$ 0.031 \\
faust\_reg & geodesic\_deterministic & 8 & 1 & 0.324 \\
faust\_reg & euclidean\_deterministic & 16 & 1 & 0.279 \\
faust\_reg & euclidean\_random & 16 & 5 & 0.303 $\pm$ 0.011 \\
faust\_reg & geodesic\_deterministic & 16 & 1 & 0.280 \\
faust\_reg & euclidean\_deterministic & 24 & 1 & 0.350 \\
faust\_reg & euclidean\_random & 24 & 5 & 0.353 $\pm$ 0.015 \\
faust\_reg & geodesic\_deterministic & 24 & 1 & 0.336 \\
faust\_reg & euclidean\_deterministic & 32 & 1 & 0.359 \\
faust\_reg & euclidean\_random & 32 & 5 & \best{0.382 $\pm$ 0.015} \\
faust\_reg & geodesic\_deterministic & 32 & 1 & \second{0.368} \\
tosca & euclidean\_deterministic & 16 & 1 & 0.500 \\
tosca & euclidean\_random & 16 & 3 & 0.486 $\pm$ 0.009 \\
tosca & geodesic\_deterministic & 16 & 1 & 0.525 \\
tosca & euclidean\_deterministic & 24 & 1 & \second{0.604} \\
tosca & euclidean\_random & 24 & 3 & 0.582 $\pm$ 0.042 \\
tosca & geodesic\_deterministic & 24 & 1 & \best{0.620} \\
\bottomrule
\end{tabular}
\end{table}

\paragraph{Information and symmetry diagnostics.}
The moment-compression diagnostic samples vertices from FAUST-Reg and TOSCA, forms sorted seed-response distributions per scale, and compares the six moment channels against two references: PCA variance captured by six components and held-out ridge reconstruction $R^2$ from moments to sorted responses.  The soft-Voronoi diagnostic reports the entropy and top1--top2 seed-assignment margin of $p_{s,t,\tau}$ at $\tau=0.01$.  The side-information diagnostic fits a logistic classifier that predicts the sign of a centered left/right coordinate on FAUST-Reg local descriptors after dropping vertices near the median plane.  It is a controlled proxy for symmetry-informative content, not a shape-correspondence benchmark.

\section{Results and Discussion}
We present results in order of the protocol cascade: retrieval (the end-to-end score), aggregation and codebook effects (the layers directly beneath it), robustness and ablations (which components contribute), and diagnostics (what the failures reveal).  The final two subsections---negative evidence as design signal, and matching diagnostics---are where the audit's methodological contribution is most visible.

\subsection{Retrieval}
\begin{figure}[t]
    \centering
    \includegraphics[width=\linewidth]{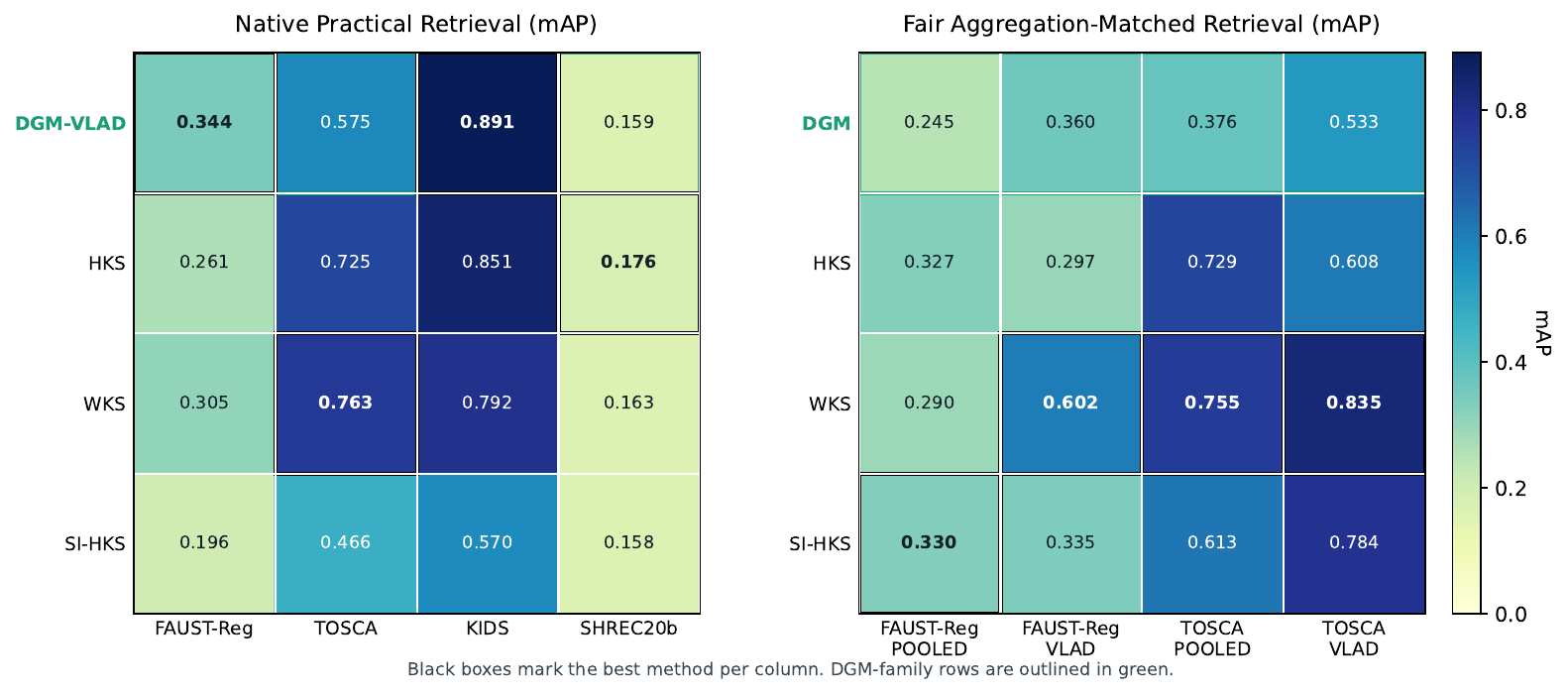}
    \caption{Native practical and aggregation-matched retrieval tell different stories.  This is the main protocol insight: global retrieval performance is a property of the descriptor plus the aggregation layer, not the local descriptor alone.}
    \label{fig:native-fair}
\end{figure}

Table~\ref{tab:native} is favorable to DGM-VLAD in several settings, including FAUST-Reg, Kids, and some native implementation metrics.  The aggregation-matched tables give a more conservative view of the local descriptor signal, and Figure~\ref{fig:native-fair} visualizes this gap between native and fair protocol readings.  In the four-method matched protocol, WKS-VLAD is much stronger than DGM-VLAD on both FAUST-Reg and TOSCA.  In the extended protocol, GMSD-HKS obtains the highest mAP/top-1 in both datasets in this implementation, with $0.621/0.820$ on FAUST-Reg and $0.865/0.963$ on TOSCA.  These results support a bounded conclusion: DGM is a seed-field moment baseline that avoids spectral decomposition and exposes a different input-signal choice, while spectral moment descriptors remain stronger in the present fair retrieval setup.

The heat-kernel approximation baselines are useful negative results.  HKS-Pad\'e and HKS-MR proxy are better than HKS-Cheb, but all three remain below DGM-VLAD and far below WKS/GMSD-HKS under VLAD-cosine.  This result does not address optimized heat-kernel approximation methods; it shows that simple drop-in approximations in the audited pipeline do not close the retrieval gap.

\FloatBarrier

\subsection{Aggregation and Codebook Transfer}
The VLAD layer is not neutral.  It improves the practical DGM pipeline on several datasets, but its codebook is dataset-dependent.  Appendix Table~\ref{tab:codebook-transfer} shows that transferring a DGM-VLAD codebook learned on FAUST-Reg to TOSCA and Kids reduces mAP.  This supports the native-vs-fair split used throughout the paper: a retrieval score reflects both the local descriptor signal and the aggregation protocol.

Appendix Table~\ref{tab:split-codebook-protocol} gives a second check on the same issue by comparing all-shapes fitted and split-fitted projection/codebook protocols.  The absolute values move under this split-aware protocol, especially on TOSCA, but the main ranking pattern remains stable: WKS and SI-HKS remain strong under VLAD, while DGM's value is tied to the particular aggregation and fitting protocol.  This reinforces the audit principle that codebook fitting is part of the measured method rather than a harmless preprocessing detail.

The repeat experiment in Table~\ref{tab:vlad-repeat} refits the VLAD codebook with five initialization seeds.  The WKS and GMSD-HKS advantage over DGM is stable on both FAUST-Reg and TOSCA, while the DGM--HKS ordering on TOSCA is not a robust separation once codebook variance is considered.

\begin{table}[!htbp]
\centering
\small
\caption{VLAD codebook repeat stability.  mAP and top-1 are higher-is-better ($\uparrow$).  Each row refits the VLAD codebook with five different initialization seeds while keeping descriptors, PCA dimensionality, and retrieval metric fixed.}
\label{tab:vlad-repeat}
\begin{tabular}{llccc}
\toprule
dataset & method & repeats & mAP $\uparrow$ & top-1 $\uparrow$ \\
\midrule
faust\_reg & WKS & 5 & \best{0.616 $\pm$ 0.033} & \best{0.862 $\pm$ 0.044} \\
faust\_reg & GMSD-HKS & 5 & \second{0.603 $\pm$ 0.022} & \second{0.832 $\pm$ 0.013} \\
faust\_reg & DGM & 5 & 0.349 $\pm$ 0.009 & 0.488 $\pm$ 0.045 \\
faust\_reg & HKS & 5 & 0.285 $\pm$ 0.028 & 0.414 $\pm$ 0.067 \\
tosca & GMSD-HKS & 5 & \best{0.869 $\pm$ 0.003} & \best{0.985 $\pm$ 0.006} \\
tosca & WKS & 5 & \second{0.845 $\pm$ 0.008} & \best{0.985 $\pm$ 0.014} \\
tosca & HKS & 5 & 0.582 $\pm$ 0.054 & 0.745 $\pm$ 0.056 \\
tosca & DGM & 5 & 0.562 $\pm$ 0.014 & \second{0.825 $\pm$ 0.029} \\
\bottomrule
\end{tabular}
\end{table}

Table~\ref{tab:protocol-cascade} summarizes the audit view numerically.  On FAUST-Reg, switching DGM from pooling to VLAD changes mAP by $+0.115$, while changing the local signal from DGM to GMSD-HKS under the same VLAD protocol changes mAP by $+0.261$.  On TOSCA, the corresponding effects are $+0.157$ and $+0.333$.  These deltas are larger than VLAD repeat standard deviations, so they are not explained by codebook initialization alone.  The table is the core methodological message: retrieval rankings are produced by a cascade, and the aggregation and input-field layers can be as important as the local moment formula.

\begin{table}[!htbp]
\centering
\small
\caption{Protocol cascade audit.  The final retrieval score changes when field construction, local descriptor signal, heat-step choice, aggregation, and codebook protocol are changed.  Signed mAP effects are shown in the last column; positive and negative effects are colored by sign, while codebook repeat reports standard deviation rather than a signed delta.}
\label{tab:protocol-cascade}
\begin{tabular}{llc}
\toprule
dataset & cascade factor & mAP effect \\
\midrule
faust\_reg & aggregation: VLAD - pooling & \textcolor{green!45!black}{+0.115} \\
faust\_reg & input signal: GMSD-HKS - DGM & \textcolor{green!45!black}{+0.261} \\
faust\_reg & heat steps: best - one-step & +0.000 \\
faust\_reg & codebook repeat std. & 0.009 \\
tosca & aggregation: VLAD - pooling & \textcolor{green!45!black}{+0.157} \\
tosca & input signal: GMSD-HKS - DGM & \textcolor{green!45!black}{+0.333} \\
tosca & heat steps: best - one-step & \textcolor{green!45!black}{+0.045} \\
tosca & codebook repeat std. & 0.014 \\
tosca & codebook transfer drop & \textcolor{red!60!black}{-0.068} \\
\bottomrule
\end{tabular}
\end{table}

\FloatBarrier

\subsection{Robustness}
\begin{figure}[t]
    \centering
    \includegraphics[width=\linewidth]{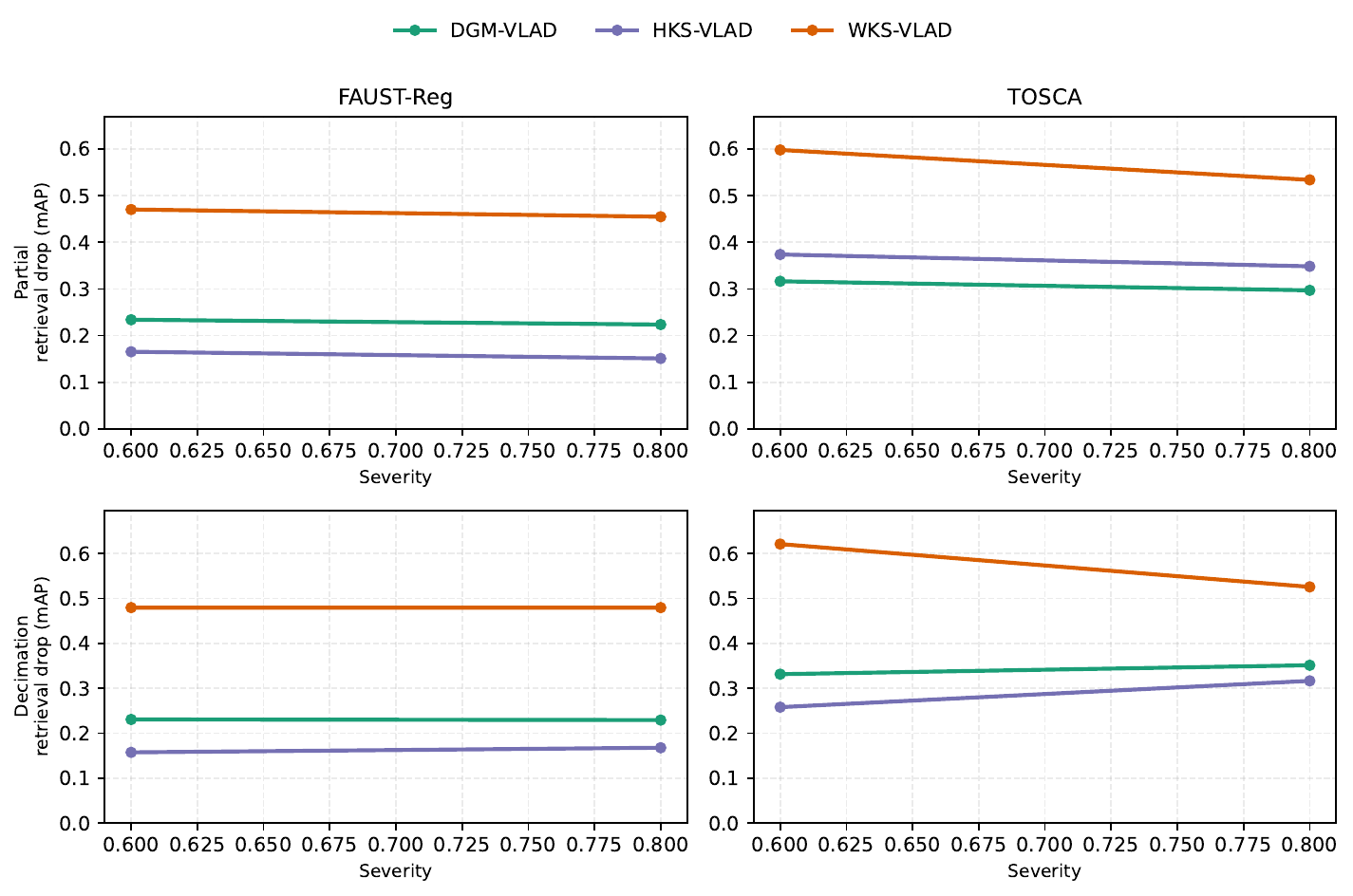}
    \caption{Frozen-codebook robustness curves.  The mixed ranking across perturbation types shows that robustness is protocol- and dataset-dependent rather than a universal property of one descriptor.}
    \label{fig:robustness-curves}
\end{figure}

Table~\ref{tab:robustness-fixed} discourages overclaiming.  On FAUST-Reg, HKS-VLAD drops less than DGM-VLAD under partiality, decimation, and noise, while WKS-VLAD is brittle in the same rows but robust under remeshing.  On TOSCA, DGM-VLAD is stronger than HKS-VLAD under partiality, remeshing, and noise, but weaker under decimation.  Figure~\ref{fig:robustness-curves} shows the same point across perturbation curves.  The TOSCA noise improvement for DGM-VLAD is best interpreted as a protocol-specific regularization effect rather than as evidence that noise generally improves the descriptor.

Appendix Table~\ref{tab:robustness-full-grid} lists the non-clean severity grid behind this summary.  The fixed-severity table is therefore not a cherry-picked slice; it is a compact view of a broader pattern in which robustness depends jointly on dataset, perturbation type, and descriptor family.

\FloatBarrier

\subsection{Field and Moment Ablations}
The field/time ablation in Appendix Table~\ref{tab:field-time} shows that the best variant is dataset-dependent.  Fixed heat scales work best on FAUST-Reg, while adaptive spectral time scales are stronger on TOSCA.  Graph-geodesic and heat-method geodesic fields are useful on TOSCA but poor on FAUST-Reg in the tested configuration.  Appendix Table~\ref{tab:field-mode-heat-method} isolates the heat-method geodesic field in the same moment-and-aggregation pipeline.  The heat-method row is particularly informative: replacing the default implicit response by a more direct heat-flow geodesic distance does not automatically improve retrieval.

\begin{figure}[t]
    \centering
    \includegraphics[width=\linewidth]{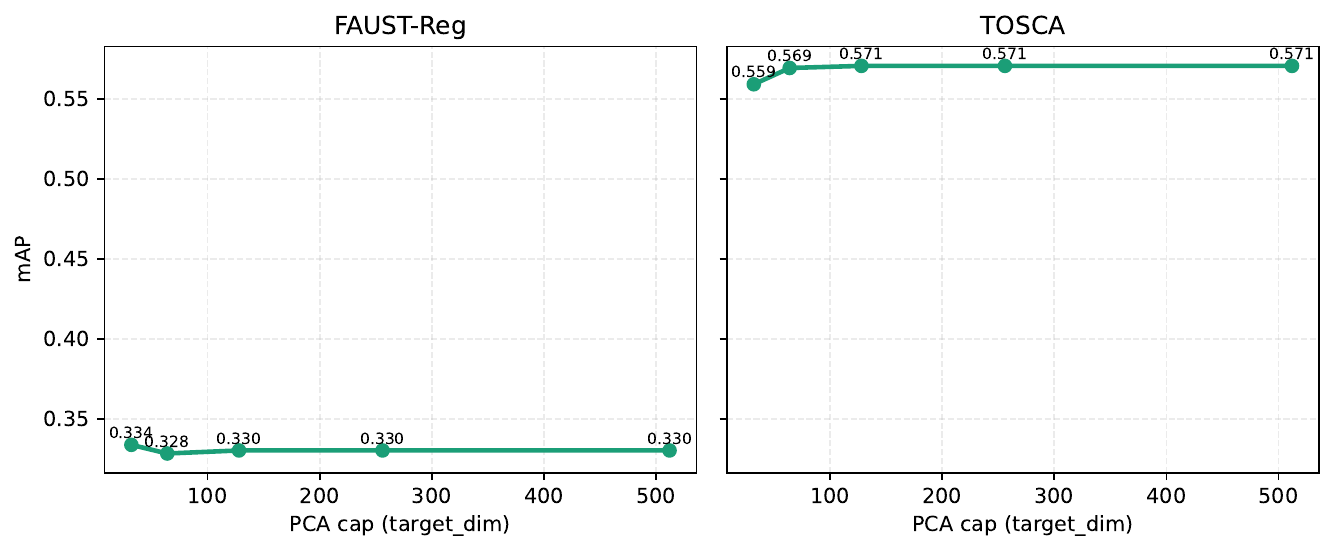}
    \caption{Dimensionality sensitivity for the fair retrieval pipeline.  The curve is included as a protocol check: projection and aggregation settings can move absolute scores, so conclusions are drawn from matched protocols rather than isolated native settings.}
    \label{fig:dimensionality}
\end{figure}

Figure~\ref{fig:dimensionality} provides a separate protocol check on projection dimensionality, showing that absolute retrieval scores can move with the dimensionality cap even when descriptor families and metrics are fixed.

Appendix Table~\ref{tab:heat-response-diagnostics} reports the raw response sign before Eq.~\ref{eq:proxy}.  The negative fractions are large, so positive-part clipping is not a cosmetic numerical epsilon; it is part of making cotangent point-source responses into a usable nonnegative field for the log transform.

The multistep diffusion ablation in Appendix Table~\ref{tab:heat-steps} addresses whether the one-step response should be replaced by sequential implicit smoothing.  This changes the finite-time field while leaving the moment and aggregation pipeline fixed.  The answer is not monotonic: FAUST-Reg prefers the one-step field, while TOSCA improves at two steps and then degrades sharply at four and eight steps.  More sequential smoothing is therefore not automatically better for DGM retrieval; the useful signal appears to come from a finite-time response with enough contrast, not from the smoothest available heat evolution.

Table~\ref{tab:dgm-linear} directly tests whether DGM's hand-designed nonlinear normalization can be replaced by a simple learned linear transform.  The result is mixed and informative.  The default nonlinear normalization is better for pooled descriptors on both FAUST-Reg and TOSCA, but PCA whitening over raw moment channels is better once VLAD is used.  Signed-log compression is therefore an aggregation-dependent engineering choice, not an intrinsically optimal descriptor theorem.  This sharpens the paper's main point: DGM is a seed-field signal family and audit tool, rather than a claim that one fixed normalization is final.

\begin{table}[!htbp]
\centering
\small
\caption{Hand nonlinear DGM normalization versus a simple learned linear whitening substitute.  All retrieval metrics are higher-is-better ($\uparrow$).  DGM-NL denotes the nonlinear default variant with z-score plus signed-log local normalization.}
\label{tab:dgm-linear}
\begin{tabular}{llcccc}
\toprule
dataset & variant & pooled mAP $\uparrow$ & pooled top-1 $\uparrow$ & VLAD mAP $\uparrow$ & VLAD top-1 $\uparrow$ \\
\midrule
faust\_reg & DGM-Linear & \second{0.160} & \second{0.230} & \best{0.375} & \second{0.480} \\
faust\_reg & DGM-NL & \best{0.245} & \best{0.380} & \second{0.360} & \best{0.530} \\
tosca & DGM-Linear & \second{0.288} & \second{0.412} & \best{0.648} & \best{0.800} \\
tosca & DGM-NL & \best{0.376} & \best{0.562} & \second{0.533} & \second{0.750} \\
\bottomrule
\end{tabular}
\end{table}

\FloatBarrier

Moment ablations in Appendix Table~\ref{tab:moment-tensor} show that mean/variance alone is insufficient for the strongest native DGM-VLAD result, while adding min/max often helps more than skew/kurtosis alone.  The tensor branch does not consistently improve the pooled DGM baseline and should be treated as optional extrinsic information rather than a core intrinsic descriptor component.

\subsection{Information Compression and Soft Voronoi Behavior}
Appendix Table~\ref{tab:information-compression} tests whether the six moment channels are a severe bottleneck.  Across FAUST-Reg and TOSCA, six PCA components explain roughly $0.80$--$0.84$ of the variance of the sorted 24-seed response distribution.  The actual DGM moments linearly reconstruct held-out sorted responses with $R^2$ around $0.58$--$0.73$.  The moments therefore retain substantial distributional structure, but they are not lossless and should not be described as sufficient statistics for the seed fields.

The same table supports the soft-Voronoi interpretation.  As the diffusion scale increases, entropy rises and the top1--top2 seed margin falls on both datasets.  FAUST-Reg moves from entropy $0.355$ and margin $0.399$ at $t=0.01$ to entropy $0.666$ and margin $0.024$ at $t=0.15$; TOSCA shows the same trend.  DGM therefore samples a continuum from near hard seed ownership to diffuse multi-seed context.  This gives the scale set a concrete role: it is not just a hyperparameter sweep, but a way to mix local Voronoi-like detail with global context.

Appendix Table~\ref{tab:ph-diagnostic} adds a small topology diagnostic using DGM seed fields as scalar filtrations.  This diagnostic is not used for retrieval.  It checks the multiscale interpretation: total zero-dimensional (0D) persistence drops sharply from $t=0.01$ to $t=0.15$ in the mean and variance seed-field channels across FAUST-Reg, TOSCA, and SMAL-toy, a local synthetic-animal sanity-check set.  The result is consistent with the view that short-time fields contain many local basins while longer-time fields smooth them into coarser context.

\subsection{Seed Stability}
Seed stability is central to any seed-conditioned descriptor.  Such descriptors necessarily break perfect intrinsic symmetry.  On a sphere, a descriptor that depends on selected seeds will assign different vectors to vertices even though the vertices are intrinsically equivalent.  This is not an implementation artifact; it is a consequence of the design.  The practical question is therefore whether the retrieval code is stable under reasonable seed choices and mesh vertex reorderings.  The deterministic and graph-geodesic seed modes make this behavior measurable.

Table~\ref{tab:seed-stability} shows two useful patterns.  First, increasing seed count helps DGM-VLAD on both datasets in the tested range: FAUST-Reg random Euclidean improves from $0.253\pm0.031$ at $k=8$ to $0.382\pm0.015$ at $k=32$, while TOSCA improves from $0.486\pm0.009$ at $k=16$ to $0.582\pm0.042$ at $k=24$.  Second, graph-geodesic deterministic FPS is not uniformly best, but it is a strong stable default: it is best on TOSCA at both tested counts and competitive on FAUST-Reg.

The permutation check in Appendix Table~\ref{tab:seed-permutation} is sharper.  Deterministic Euclidean and graph-geodesic FPS produce essentially identical descriptors after vertex reordering: global cosine is numerically one and aligned local relative error is near $10^{-11}$ on both datasets.  Random Euclidean FPS is not vertex-order stable in this implementation, with local relative error around $0.56$ on FAUST-Reg and $0.78$ on TOSCA.  Deterministic seeding is therefore the appropriate default, while random FPS is best interpreted as a variance ablation.

\subsection{Symmetry-Informative Side Diagnostic}
Purely intrinsic signatures can be intentionally symmetry-agnostic: symmetric points receive the same response unless external constraints disambiguate them.  Recent representation work makes a similar distinction between symmetry-informative and symmetry-agnostic features \cite{weissberg2026symmetry}.  DGM sits on the informative side when a deterministic seed frame is used.  Appendix Table~\ref{tab:symmetry-side} quantifies this behavior with a left/right side proxy on FAUST-Reg.  DGM local descriptors reach $0.836\pm0.057$ balanced accuracy, while HKS, WKS, and GMSD-HKS stay near chance.  This does not mean DGM is a strong correspondence method; it means that seed conditioning injects side information that diagonal spectral descriptors deliberately suppress.

A seed-count sweep makes this point sharper.  With Euclidean seed frames, side information is already strong at $k=4$ and decreases mildly as more seeds average the frame: deterministic Euclidean sampling gives $0.868$ balanced accuracy at $k=4$ and $0.830$ at $k=48$, while five random Euclidean repeats give $0.887$ at $k=4$ and $0.823$ at $k=48$.  Graph-geodesic deterministic seeds behave differently, rising from $0.650$ at $k=4$ to about $0.81$ by $k=16$.  Thus symmetry informativeness is not a monotone function of seed count.  It is a controllable property of the seed frame and should be selected according to the downstream task.

\subsection{Foundation-Feature Probe}
Appendix Table~\ref{tab:foundation-probe} tests a zero-shot alternative inspired by lifted foundation-feature pipelines: render each FAUST-Reg mesh from eight views, extract DINOv2-small patch tokens, lift them back to vertices, and pool them into global descriptors.  DINOv2 improves over rendered normal features, but it remains below the geometry-specific descriptors in this simple lifting protocol.  The result is a bounded negative result: a modern pretrained image model does not automatically solve this geometry retrieval problem without a stronger 3D lifting and aggregation design.

\subsection{Descriptor--Solver Compatibility}
The Python Functional Maps (pyFM) and ZoomOut diagnostic in Table~\ref{tab:pyfm-correspondence} inserts DGM, HKS, and WKS into a functional-map matching pipeline on FAUST-Reg same-subject pairs.  DGM has the lowest nearest-neighbor error in this check, but WKS is substantially better after functional-map optimization and ZoomOut refinement.  The mechanism is structural: DGM descriptors are conditioned on independently selected seed frames, while the functional-map objective assumes that descriptor coordinates are comparable functions whose preservation can be enforced linearly in a shared spectral basis.  Independent seed frames can preserve useful nearest-neighbor neighborhoods but still violate this descriptor-preservation assumption.  Matching is therefore reported as a solver-compatibility diagnostic rather than as an application claim.

\begin{table}[!htbp]
\centering
\small
\caption{pyFM and ZoomOut correspondence diagnostic on 20 FAUST-Reg same-subject pairs.  Geodesic error is lower-is-better ($\downarrow$), while hit@10\% is higher-is-better ($\uparrow$).  DGM NN descriptors are competitive, but the standard functional-map optimization favors WKS in this setting.}
\label{tab:pyfm-correspondence}
\begin{tabular}{lcccc}
\toprule
method & NN geo. err. $\downarrow$ & NN hit@10\% $\uparrow$ & pyFM geo. err. $\downarrow$ & pyFM hit@10\% $\uparrow$ \\
\midrule
WKS & \second{1.155} & \second{0.216} & \best{0.754} & \best{0.381} \\
HKS & 1.209 & 0.203 & \second{0.864} & \second{0.339} \\
DGM & \best{0.782} & \best{0.450} & 1.205 & 0.066 \\
\bottomrule
\end{tabular}
\end{table}

\FloatBarrier

\subsection{Negative Evidence as Design Signal}
Several experiments were originally natural attempts to improve DGM's retrieval standing.  They did not make DGM the strongest descriptor, but they revealed failure modes whose structure is informative for the field.  Rather than treating these as abandoned attempts, we organize them as design probes---controlled experiments that turn a ranking shortfall into a testable statement about descriptor construction, solver compatibility, or discretization.  Appendix Table~\ref{tab:mechanism-probes} lists the probes and their interpretations; Appendix Table~\ref{tab:negative-insights} records the design lessons extracted from each plausible shortcut that the audit tested and rejected.

The first mechanism probe tests a tempting explanation: any perturbation might make an exact heat-method field more useful.  The evidence is limited.  Adding controlled Gaussian perturbations to the heat-method geodesic fields improves FAUST-Reg VLAD mAP from $0.125$ to $0.133$ and TOSCA VLAD mAP from $0.534$ to $0.555$, with TOSCA pooled descriptors also improving from $0.505$ to $0.547$.  The gains are real but small compared with the gap between implicit DGM and the heat-method field on FAUST-Reg.  The useful error is therefore not arbitrary noise; it is structured roughness induced by the finite-step cotangent resolvent, the log transform, and the aggregation protocol.

The raw-response diagnostic clarifies Eq.~\ref{eq:proxy}.  Before clipping and logarithmic normalization, the implicit cotangent response is negative for roughly half of the vertex--seed--scale samples on FAUST-Reg and TOSCA.  These negative values are not rare floating-point noise, and their vertexwise occurrence correlates only weakly with simple mesh-quality proxies such as valence deviation, edge-length coefficient of variation, face aspect ratio, and vertex area.  The practical DGM field is therefore better described as a signed cotangent-resolvent signal that is converted into a distance-like nonnegative field, not as a positivity-preserving heat kernel evaluation.

The learned-linear and foundation-feature probes tell a different story.  A simple PCA whitening substitute improves DGM under VLAD but not under pooling, so hand-designed nonlinear normalization should be treated as part of an aggregation-dependent block rather than as an isolated descriptor theorem.  Meanwhile, a straightforward DINOv2 lifting pipeline underperforms geometry-specific descriptors on full FAUST-Reg retrieval.  This does not weaken foundation features in general; it indicates that view-based semantic features need a geometry-aware lifting and aggregation mechanism before they can replace surface-native signals in this setting.

Table~\ref{tab:followup-design-probes} collects three follow-up probes suggested by the negative evidence; the table appears immediately below.

\begin{table}[!htbp]
\centering
\small
\caption{Follow-up design probes motivated by the failure analysis.  Values are mAP ($\uparrow$).  The probes separate three possible responses to DGM's limitations: adding random roughness, concatenating signals before a shared codebook, and fusing scores after each descriptor has its own aggregation.}
\label{tab:followup-design-probes}
\begin{tabular}{p{0.24\linewidth}p{0.20\linewidth}p{0.20\linewidth}p{0.25\linewidth}}
\toprule
probe & FAUST-Reg mAP $\uparrow$ & TOSCA mAP $\uparrow$ & interpretation \\
\midrule
Raw-response DGM noise, VLAD &
0.361 clean; \best{0.377} best &
0.583 clean; \best{0.597} best &
Small gains only; random noise is not the main source of useful roughness. \\
Local DGM+GMSD-HKS+WKS, VLAD &
\best{0.697} spectral; 0.425 with DGM &
\best{0.864} spectral; 0.818 with DGM &
Naive local concatenation lets weaker high-dimensional channels dilute the shared codebook. \\
Late distance fusion with DGM &
0.662 base; \best{0.673} best &
0.857 base; \best{0.862} best &
DGM is more useful as weak auxiliary evidence than as an equal local codebook input. \\
\bottomrule
\end{tabular}
\end{table}

\paragraph{Structured roughness.}
Injecting noise into the raw implicit response before clipping and logarithmic conversion gives only small VLAD gains, from $0.361$ to $0.377$ on FAUST-Reg and from $0.583$ to $0.597$ on TOSCA.  Together with the heat-method noise probe in Table~\ref{tab:mechanism-probes}, this supports a precise conclusion: useful roughness is structured by the cotangent resolvent, finite diffusion scale, log transform, and aggregation protocol; it is not reproduced by arbitrary Gaussian perturbation.

\paragraph{Protocol-aware fusion.}
Naive local concatenation is a poor way to combine DGM with stronger spectral signals: adding DGM to a GMSD-HKS+WKS local stack substantially hurts a shared VLAD codebook.  Late distance fusion is more sensible.  A small DGM weight improves a strong spectral distance ensemble slightly on both datasets, while large DGM weights degrade it.  DGM therefore behaves like weak complementary evidence, not like an equal replacement for the main spectral signal.  The lesson is broader than DGM: heterogeneous local channels should not be forced into one codebook before checking whether their scales, coordinate systems, and residual statistics are compatible.

The matching diagnostic is the clearest cautionary result.  DGM gives the best nearest-neighbor correspondence signal but becomes the weakest descriptor after functional-map optimization.  The failure is informative because it separates local discriminability from solver compatibility.  A descriptor can be locally meaningful while still being poorly expressed as comparable functions in two independently chosen spectral bases.  Seed-conditioned descriptors therefore require synchronized seeds, cross-shape probe alignment, or a matching objective designed around seed-frame ambiguity before they can be used as standard functional-map preservation terms.

\begin{table}[!htbp]
\centering
\small
\caption{Spectral compressibility diagnostic on 20 FAUST-Reg meshes.  $R^2$ columns are higher-is-better ($\uparrow$) as retained-energy measures.  Roughness is descriptive and is not ranked.}
\label{tab:spectral-compressibility}
\begin{tabular}{lrrrr}
\toprule
method & roughness & $R^2_{32}$ $\uparrow$ & $R^2_{64}$ $\uparrow$ & $R^2_{96}$ $\uparrow$ \\
\midrule
HKS & 660.2 & \best{0.165} & \best{0.323} & 0.772 \\
WKS & 357.5 & \second{0.088} & \second{0.174} & \second{0.877} \\
GMSD-HKS & 7944.1 & 0.059 & 0.110 & 0.699 \\
DGM & 968.2 & 0.026 & 0.101 & \best{0.891} \\
\bottomrule
\end{tabular}
\end{table}

Table~\ref{tab:spectral-compressibility} tests the basis side of this explanation.  Under the ground-truth registered FAUST topology, we project each local descriptor channel onto the first Laplacian eigenfunctions of the same mesh and measure retained mass-weighted energy.  DGM retains only $2.6\%$ of its energy in the first 32 modes and $10.1\%$ in the first 64 modes, below HKS and WKS in the range commonly used by functional-map solvers.  At 96 modes, however, DGM reaches $89.1\%$ retained energy.  This pattern reframes the failed functional-map result: DGM is not simply missing information; much of its discriminative content lies outside the low-frequency subspace where the solver expects descriptors to live.

\begin{table}[!htbp]
\centering
\small
\caption{DGM bandwidth diagnostic on 20 FAUST-Reg same-subject pairs.  Error columns are lower-is-better ($\downarrow$); hit@10\% and CSAS are higher-is-better ($\uparrow$).  The unprojected DGM nearest-neighbor baseline is 0.782 mean geodesic error and 0.450 hit@10\% diameter.}
\label{tab:pyfm-bandwidth}
\begin{tabular}{@{}cccccc@{}}
\toprule
$k$ & spec. NN err. $\downarrow$ & spec. NN hit@10\% $\uparrow$ & spec. CSAS $\uparrow$ & pyFM err. $\downarrow$ & pyFM hit@10\% $\uparrow$ \\
\midrule
32 & 1.089 & 0.118 & \second{0.743} & 1.169 & 0.148 \\
64 & 1.080 & 0.163 & 0.668 & 1.081 & 0.221 \\
96 & \second{0.996} & \second{0.280} & 0.669 & \second{1.056} & \second{0.243} \\
128 & \best{0.788} & \best{0.446} & \best{0.835} & \best{0.992} & \best{0.264} \\
\bottomrule
\end{tabular}
\end{table}

Table~\ref{tab:pyfm-bandwidth} tests whether this observation is enough to repair the failure inside pyFM/ZoomOut.  Increasing the basis improves DGM after functional-map refinement: the pyFM error drops from $1.169$ at 32 modes to $0.992$ at 128 modes, and hit@10\% diameter rises from $0.148$ to $0.264$.  The projected nearest-neighbor descriptor also recovers the original DGM signal by 128 modes.  The repair is incomplete, however: the unprojected DGM nearest-neighbor error remains lower ($0.782$), and WKS in Table~\ref{tab:pyfm-correspondence} remains stronger after standard refinement.  Thus bandwidth mismatch is a real part of the failure, but not the whole explanation.  Seed-conditioned descriptors also need cross-shape seed alignment or a matching objective that does not assume comparable descriptor coordinates.

\begin{table}[!htbp]
\centering
\small
\caption{Ground-truth seed synchronization diagnostic on 20 FAUST-Reg same-subject pairs at 128 functional-map modes.  Overlap, CSAS, and hit@10\% are higher-is-better ($\uparrow$); error columns are lower-is-better ($\downarrow$).}
\label{tab:synchronized-seed-dgm}
\begin{tabular}{llccccc}
\toprule
seed frame & policy & overlap $\uparrow$ & CSAS $\uparrow$ & NN err. $\downarrow$ & pyFM err. $\downarrow$ & pyFM hit@10\% $\uparrow$ \\
\midrule
Euclidean & independent & 0.071 & 0.740 & 0.750 & 1.010 & 0.248 \\
Euclidean & synchronized & \best{1.000} & \best{0.831} & \second{0.607} & \best{0.834} & \second{0.375} \\
Geodesic & independent & \second{0.179} & 0.737 & 0.784 & 1.024 & 0.227 \\
Geodesic & synchronized & \best{1.000} & \second{0.827} & \best{0.447} & \second{0.848} & \best{0.381} \\
\bottomrule
\end{tabular}
\end{table}

Table~\ref{tab:synchronized-seed-dgm} tests this seed-alignment hypothesis directly as an upper bound.  On registered FAUST-Reg pairs, we select deterministic seeds on the source shape and copy the seed indices to the target through the known correspondence before recomputing DGM.  This improves cross-shape descriptor alignment and matching.  For geodesic deterministic seeds, NN error drops from $0.784$ to $0.447$, and pyFM error drops from $1.024$ to $0.848$.  The result is positive but bounded: synchronized DGM is much better than independently seeded DGM, yet it still does not surpass the WKS pyFM result in Table~\ref{tab:pyfm-correspondence}.  Seed synchronization is therefore a real missing component for seed-conditioned descriptors, but it is not by itself a complete matching pipeline.

The upper-bound experiment also defines a concrete path toward an unsupervised version.  A deployable synchronized-seed descriptor could first compute a coarse seed-agnostic map using a high-CSAS signal such as WKS, HKS, or GMSD-WKS, use that map only to transfer an initial seed frame, and then refine the seed assignment by matching seed-response signatures under a coverage regularizer.  The synchronized seed frame would then be used to recompute seed-conditioned fields and to optimize a matching objective that combines descriptor preservation with diffusion consistency.  The evaluation criteria are explicit: an unsupervised method should be compared with the ground-truth synchronized upper bound, report seed overlap or assignment consistency, improve CSAS, and improve the final matching solver rather than only the nearest-neighbor descriptor.  This route keeps the present paper as an audit while turning its strongest correspondence diagnostic into a well-specified follow-up method.

A complementary diagnostic reaches the same conclusion without relying on a particular solver.  CSAS fits a ridge-linear map from source descriptor coordinates to target descriptor coordinates on ground-truth corresponding vertices and reports held-out $R^2$.  High CSAS indicates that descriptor channels behave like comparable probe functions across shapes.

\begin{table}[!htbp]
\centering
\small
\caption{Expanded CSAS diagnostic on 20 FAUST-Reg same-subject pairs.  Direct cosine, CSAS, and ground-truth nearest-neighbor accuracy are higher-is-better ($\uparrow$); relative residual is lower-is-better ($\downarrow$).}
\label{tab:csas-extended}
\begin{tabular}{lcccc}
\toprule
descriptor & direct cosine $\uparrow$ & linear CSAS $R^2$ $\uparrow$ & rel. residual $\downarrow$ & GT nearest-neighbor acc. $\uparrow$ \\
\midrule
WKS & \best{0.989} & \best{0.982} & \best{0.123} & 0.012 \\
HKS & 0.939 & \second{0.965} & \second{0.172} & 0.001 \\
GMSD-WKS & \second{0.968} & 0.960 & 0.185 & \best{0.075} \\
SI-HKS & 0.885 & 0.942 & 0.231 & 0.004 \\
GMSD-HKS & 0.727 & 0.814 & 0.431 & 0.036 \\
DGM & 0.601 & 0.755 & 0.494 & 0.011 \\
DINOv2 lift & 0.506 & 0.510 & 0.700 & \second{0.052} \\
normal lift & 0.586 & 0.353 & 0.804 & 0.007 \\
\bottomrule
\end{tabular}
\end{table}

Table~\ref{tab:csas-extended} expands this diagnostic beyond the three descriptors used in the first matching table and includes the simple lifted foundation-feature probes from Appendix Table~\ref{tab:foundation-probe}.  WKS, HKS, GMSD-WKS, and SI-HKS all have high coordinate alignment; GMSD-HKS is intermediate; DGM is lower; the DINOv2 and rendered-normal lifts are lower still under this projection protocol.  This ordering differs from retrieval accuracy and nearest-neighbor correspondence quality, which is precisely why CSAS is useful.  It tests an interface assumption of functional-map solvers: descriptor coordinates should be comparable across shapes before they are enforced as preserved functions.  DGM violates this assumption more strongly than the spectral families, while synchronized seeds in Table~\ref{tab:synchronized-seed-dgm} partially repair it.  Simple lifted image features do not automatically satisfy the same contract either.  In this sense, CSAS operationalizes a descriptor--solver contract that is often implicit in classical matching pipelines.

Finally, a decimation sanity check prevents an overinterpretation of the FAUST-Reg result.  The original FAUST-Reg meshes share one topology, whereas TOSCA contains varied vertex and face counts.  However, after $50\%$ simplification with tiny random geometric jitter, the FAUST meshes have 100 distinct topology hashes and implicit DGM-VLAD still remains much stronger than the heat-method field ($0.344$ versus $0.136$ mAP).  Shared tessellation can amplify protocol effects, but it is not the sole explanation for the implicit-field advantage.
\FloatBarrier

\section{Design Recommendations}
The audit suggests a simple reporting rule: native and aggregation-matched scores should both be shown.  Native pipelines measure end-to-end utility, while matched pipelines separate the local signal from the aggregation layer.  When the two protocols disagree on method rankings, the disagreement is itself evidence about where performance is coming from.

The codebook should be treated as a fitted model.  VLAD or bag-of-features results should report the fitting split, initialization repeat variance, and any transfer protocol.  A single codebook fit can hide a source of variance large enough to change the ordering of methods whose local signals are close.

For descriptor design, the first question should be the input field, not the moment formula.  In these experiments, the field construction layer has a larger effect on final accuracy than adding more moments or changing the normalization.  GMSD-HKS is the accuracy reference in the reported fair retrieval protocol.  DGM is most useful when sparse solves, symmetry-informative side information, or non-spectral deployment are the priority.  When DGM is combined with stronger spectral descriptors, late score-level fusion is safer than forcing all local channels through one shared codebook.

Finally, deterministic seed rules should be the default whenever a seed-conditioned descriptor is reported.  Random farthest-point sampling is useful for variance analysis, but the default descriptor should be vertex-order independent.  For shape matching, nearest-neighbor descriptor quality is not sufficient evidence: seed-conditioned features need seed alignment across shapes, synchronized probe fields, or a dedicated matching objective before they become competitive inputs for standard functional-map solvers.

\section{Reproducibility and Release Package}
The experiments are designed to be reproducible from a release package rather than implicit local state.  Each reported run is associated with a configuration, a fixed random seed unless seed variance is explicitly studied, cached local and global descriptors, and the metadata needed to reconstruct the evaluated shape order.  The retrieval, robustness, timing, seed-stability, heat-response, multistep-diffusion, field-mode, moment-compression, foundation-feature, functional-map bandwidth, seed-synchronization, hybrid-fusion, and persistence diagnostics are implemented as separate reproducible experiment modules so that individual tables can be regenerated without rerunning the entire study.  The GMSD rows are independent implementations from the published description, not the authors' released code; they are reported as controlled baselines rather than as bitwise reproductions of the original implementation.  The heat-approximation rows are controlled implementations in this repository, not optimized reference implementations of the full approximation literature.  The test suite covers Laplacian assembly, moment descriptors, sampling, perturbations, optional heat-method fields, and the extended baselines.

\paragraph{Code availability.}
The implementation, preprocessing pipeline, experiment configurations, table and figure generation code, and tests will be made publicly available after acceptance.  The public release will include an explicit code license, the exact configurations, and run metadata used to reproduce the reported tables and figures.  For journal publication, the code package can be frozen as an archival release with a digital object identifier (DOI) so that the reported numbers can be reproduced independently from the development repository.

\paragraph{Data availability.}
The experiments use public benchmark datasets obtained from their original providers: FAUST from the FAUST project website (\url{http://faust.is.tue.mpg.de}), TOSCA from the public TOSCA high-resolution resources (\url{http://tosca.cs.technion.ac.il/book/resources_data.html}), Kids from the TUM Computer Vision Group dataset page (\url{https://cvg.cit.tum.de/data/datasets/kids}), and SHREC20B from the SHREC'20 non-isometric correspondence benchmark page (\url{http://robertodyke.com/shrec2020/}).  SCAPE preprocessing manifests are included in the planned code release for correspondence-oriented use, but SCAPE is not included in the class-retrieval tables because the processed set has one retrieval label.  Raw meshes are not redistributed when the original dataset licenses or terms restrict redistribution.  The release will provide the processing routines and manifests used in this study.  Processing consists of reading the original meshes, converting them to a common Object File Format (OFF) representation when necessary, centering and area-normalizing the surface, assigning shape identifier, label, and split metadata, and writing the manifests consumed by descriptor extraction.  Synthetic robustness perturbations are generated from the processed meshes by the released perturbation routines.

\FloatBarrier
\section{Limitations}
The main limitation is novelty relative to the long history of geodesic statistics and moment descriptors.  Moment statistics over shape descriptors are established ideas; DGM's distinctive element is the seed-conditioned off-diagonal field input and the audit protocol around it, not the use of moments in isolation.

DGM is also only intrinsic-oriented.  Euclidean farthest-point sampling and tensor invariants are extrinsic.  Graph-geodesic sampling improves the situation but still breaks exact symmetries through deterministic tie breaking.  This is useful side information in some settings and a limitation in others.

The heat interpretation is similarly bounded.  Equation~\ref{eq:heat-solve} is a practical regularized implicit response, and Eq.~\ref{eq:proxy} is a normalized log field inspired by Varadhan's formula, not a finite-time geodesic distance theorem.  The timing table compares against eigendecomposition baselines in the present implementation; the supplementary heat-kernel approximation rows do not reproduce optimized reference implementations, GPU kernels, or prefactored heat-kernel alternatives.  On GPU-oriented deployments, polynomial or Chebyshev filtering and learned surface networks may define more favorable engineering points than the CPU sparse-solve regime evaluated here.

DGM also scales linearly in the number of seed-scale solves after factorization.  The reported 24-seed setting is practical for the audited meshes, but substantially denser seed frames, very high-resolution meshes, or many time scales would require batching, multigrid, prefactorization reuse, or approximate field construction.  The current robustness diagnostics use clean benchmark meshes and synthetic perturbations; non-manifold meshes, holes, disconnected components, and severe scan artifacts are outside the tested regime even though the negative-response diagnostics suggest that mesh quality can interact with the signed cotangent-resolvent field.

Finally, the retrieval audit does not establish DGM as a correspondence method.  The functional-map diagnostic shows that WKS remains stronger after standard refinement, and larger unified correspondence suites such as BeCoS \cite{ehm2025becos} require a separate benchmark protocol.  The 0D persistence diagnostic likewise supports the multiscale-field interpretation but does not turn DGM into a persistent-homology descriptor.

\section{Conclusion}
This paper has conducted a protocol-cascade audit of training-free 3D shape retrieval, using Diffused Geodesic Moments both as a practical non-spectral descriptor and as a diagnostic instrument.  The experiments lead to three principal conclusions.

First, {\em the input field and the aggregation protocol can dominate the moment formula.}  Under the matched VLAD-cosine protocol and repository implementations used here, GMSD-HKS obtains the highest scores on FAUST-Reg and TOSCA, while DGM is competitive when spectral decomposition is impractical, when symmetry-informative side information is needed, or when central processing unit (CPU)-only deployment is a constraint.  No single descriptor is best in all settings, and the choice of field and aggregation should be treated as a first-class algorithmic decision.

Second, {\em negative evidence is actionable.}  In the controlled implementations and matched protocols tested here, finite-step implicit heat responses can be more useful for retrieval than smoother heat-kernel or heat-method geodesic fields.  The effect appears to come from structured roughness introduced by the cotangent resolvent, finite diffusion scale, log transform, and aggregation protocol, not from arbitrary numerical noise.  Naive concatenation of heterogeneous local descriptors before a shared VLAD codebook can harm retrieval, while late score-level fusion preserves the contributions of each signal.  Hand-designed nonlinear descriptor normalization should be evaluated jointly with the aggregation layer, not in isolation.

Third, {\em descriptor--solver compatibility is measurable and should be audited.}  CSAS and the spectral-compressibility diagnostic introduced here provide quantitative tests of whether a descriptor satisfies the assumptions of functional-map solvers.  Across eight descriptor families spanning classical spectral signatures, geodesic-moment variants, DGM, and lifted foundation features, CSAS ranges from $0.353$ for rendered-normal lifting to $0.982$ for WKS.  This range is larger than many retrieval-score gaps between competing descriptors, confirming that solver compatibility is an independent dimension of descriptor quality.  Seed-conditioned descriptors such as DGM violate this contract more strongly than spectral families; seed synchronization and increased spectral bandwidth each repair part of the gap, but neither eliminates it entirely.  This finding generalizes beyond DGM: any descriptor with a non-intrinsic coordinate frame will face the same structural incompatibility, and future matching pipelines should either accommodate such frames or justify why they are not needed.  The synchronized-seed upper bound supplies a concrete target for that future work: replace the known correspondence by unsupervised seed transfer and assignment, then evaluate by CSAS and final solver performance rather than by seed overlap alone.

The manuscript source package is self-contained for arXiv compilation, and the planned artifact release will provide the code, configurations, and run metadata needed to reproduce every reported number and diagnostic.

\FloatBarrier
\clearpage
\bibliographystyle{plain}
\bibliography{references}

\appendix
\clearpage
\section{Supplementary Diagnostic Tables}
This appendix contains the supporting tables that are referenced from the main text but are not part of the primary narrative.  The tables are grouped by the protocol layer they test.  Main-result tables are not repeated here; the appendix is reserved for timing, transfer, split-aware codebook fitting, robustness severity grids, field construction, numerical behavior, moment compression, topology, seed invariance, symmetry, and foundation-feature probes.

\subsection{Timing and Aggregation Diagnostics}
This block reports the secondary timing, transfer, and split-aware codebook diagnostics.  They support the protocol-audit claim that extraction cost and codebook fitting are part of the measured pipeline, but they are not the headline retrieval comparison.

\begin{table}[!htbp]
\centering
\small
\caption{Average descriptor extraction time per shape for the extended baselines. Runtime columns are lower-is-better ($\downarrow$). GMSD rows include the repository's spectral HKS/WKS construction; HKS-Cheb, HKS-Pad\'e, and HKS-MR proxy avoid a full spectral basis but are simple controlled implementations.}
\label{tab:extended-timing}
\begin{tabular}{lccc}
\toprule
method & local dim & FAUST-Reg sec./shape $\downarrow$ & TOSCA sec./shape $\downarrow$ \\
\midrule
GMSD-HKS & 6 & 0.499 & 2.998 \\
GMSD-WKS & 6 & 0.596 & 2.934 \\
HKS-Cheb & 4 & \best{0.084} & \best{0.413} \\
HKS-Pad\'e & 4 & 0.302 & 2.169 \\
HKS-MR proxy & 24 & \second{0.153} & \second{0.630} \\
\bottomrule
\end{tabular}
\end{table}

\begin{table}[!htbp]
\centering
\small
\caption{DGM-VLAD codebook transfer from FAUST-Reg to other datasets. Retrieval mAP is higher-is-better ($\uparrow$); transfer gap is in-domain mAP minus FAUST-codebook mAP, so lower is better ($\downarrow$).}
\label{tab:codebook-transfer}
\begin{tabular}{lccc}
\toprule
target & FAUST-codebook mAP $\uparrow$ & in-domain mAP $\uparrow$ & transfer gap $\downarrow$ \\
\midrule
kids & \second{0.683} & \best{0.864} & 0.181 \\
tosca & \second{0.465} & \best{0.533} & 0.068 \\
\bottomrule
\end{tabular}
\end{table}

\begin{table}[!htbp]
\centering
\small
\caption{Split-aware codebook protocol check.  Retrieval mAP is higher-is-better ($\uparrow$).  The all-fit protocol fits the projection or codebook on all evaluated shapes, while the split-fit protocol fits it only on a held-out fitting split before evaluation.  The last column reports split-fit mAP minus all-fit mAP.}
\label{tab:split-codebook-protocol}
\begin{tabular}{lllccc}
\toprule
dataset & aggregation & method & all-fit mAP $\uparrow$ & split-fit mAP $\uparrow$ & $\Delta$ \\
\midrule
faust\_reg & pooled & DGM & 0.245 & 0.252 & +0.007 \\
faust\_reg & pooled & HKS & 0.327 & 0.306 & -0.021 \\
faust\_reg & pooled & SI-HKS & 0.330 & 0.366 & +0.036 \\
faust\_reg & pooled & WKS & 0.290 & 0.266 & -0.023 \\
faust\_reg & VLAD & DGM & 0.360 & 0.337 & -0.023 \\
faust\_reg & VLAD & HKS & 0.297 & 0.289 & -0.008 \\
faust\_reg & VLAD & SI-HKS & 0.335 & 0.336 & +0.001 \\
faust\_reg & VLAD & WKS & 0.602 & 0.555 & -0.047 \\
tosca & pooled & DGM & 0.376 & 0.353 & -0.023 \\
tosca & pooled & HKS & 0.729 & 0.689 & -0.041 \\
tosca & pooled & SI-HKS & 0.613 & 0.506 & -0.107 \\
tosca & pooled & WKS & 0.755 & 0.700 & -0.055 \\
tosca & VLAD & DGM & 0.533 & 0.482 & -0.051 \\
tosca & VLAD & HKS & 0.608 & 0.498 & -0.109 \\
tosca & VLAD & SI-HKS & 0.784 & 0.760 & -0.024 \\
tosca & VLAD & WKS & 0.835 & 0.793 & -0.043 \\
\bottomrule
\end{tabular}
\end{table}

\FloatBarrier
\subsection{Robustness Severity Grid}
The fixed-severity robustness table in the main text summarizes one comparable perturbation level per perturbation type.  The table below reports all non-clean severity levels used in the robustness sweep, so the fixed table can be read as a compact summary of the full diagnostic rather than as a selected outlier.

\begin{table}[!htbp]
\centering
\small
\caption{Full non-clean robustness severity grid, reported as mAP drop from the clean run ($\downarrow$).  Clean identity severities are omitted because their drop is zero by definition.  Negative values mean the perturbed run scored slightly higher than the clean run.}
\label{tab:robustness-full-grid}
\begin{tabular}{lllccc}
\toprule
dataset & perturb. & severity & DGM-VLAD drop $\downarrow$ & HKS-VLAD drop $\downarrow$ & WKS-VLAD drop $\downarrow$ \\
\midrule
faust\_reg & decimation & 0.600 & 0.231 & 0.158 & 0.480 \\
faust\_reg & decimation & 0.800 & 0.229 & 0.168 & 0.480 \\
faust\_reg & noise & 0.010 & 0.213 & 0.175 & 0.453 \\
faust\_reg & noise & 0.020 & 0.221 & 0.171 & 0.461 \\
faust\_reg & noise & 0.050 & 0.244 & 0.171 & 0.470 \\
faust\_reg & partial & 0.600 & 0.234 & 0.166 & 0.470 \\
faust\_reg & partial & 0.800 & 0.224 & 0.151 & 0.455 \\
faust\_reg & remeshing & 0.200 & 0.005 & 0.044 & -0.034 \\
faust\_reg & remeshing & 0.400 & 0.014 & 0.051 & -0.035 \\
tosca & decimation & 0.600 & 0.332 & 0.258 & 0.621 \\
tosca & decimation & 0.800 & 0.351 & 0.317 & 0.526 \\
tosca & noise & 0.010 & -0.136 & 0.101 & 0.214 \\
tosca & noise & 0.020 & -0.125 & 0.052 & 0.143 \\
tosca & noise & 0.050 & -0.019 & 0.083 & 0.151 \\
tosca & partial & 0.600 & 0.316 & 0.374 & 0.598 \\
tosca & partial & 0.800 & 0.297 & 0.349 & 0.534 \\
tosca & remeshing & 0.200 & 0.012 & 0.061 & -0.003 \\
tosca & remeshing & 0.400 & 0.040 & 0.052 & -0.009 \\
\bottomrule
\end{tabular}
\end{table}

\FloatBarrier
\subsection{Field, Heat, and Moment Diagnostics}
These tables isolate the effect of field construction, heat-response computation, and moment-channel choices while keeping the rest of the descriptor pipeline fixed.  They explain why the final descriptor should be interpreted as a finite-time, normalized response statistic rather than as a direct geodesic-distance theorem.

\begin{table}[!htbp]
\centering
\small
\caption{Field and time-scale ablation for DGM-VLAD. mAP is higher-is-better ($\uparrow$), and the best field choice is dataset-dependent.}
\label{tab:field-time}
\begin{tabular}{lllc}
\toprule
dataset & field & time scales & mAP $\uparrow$ \\
\midrule
faust\_reg & graph\_geodesic & fixed & 0.128 \\
faust\_reg & heat\_method\_geodesic & fixed & 0.125 \\
faust\_reg & heat & adaptive\_spectrum & \second{0.205} \\
faust\_reg & heat & fixed & \best{0.344} \\
tosca & graph\_geodesic & fixed & 0.515 \\
tosca & heat\_method\_geodesic & fixed & 0.555 \\
tosca & heat & adaptive\_spectrum & \best{0.668} \\
tosca & heat & fixed & \second{0.575} \\
\bottomrule
\end{tabular}
\end{table}

\begin{table}[!htbp]
\centering
\small
\caption{Heat-method geodesic field diagnostic using an external heat-method solver.  mAP and top-1 are higher-is-better ($\uparrow$); runtime is lower-is-better ($\downarrow$).  Runtime is identical within each dataset because pooling and VLAD share the same field extraction.}
\label{tab:field-mode-heat-method}
\begin{tabular}{lllccc}
\toprule
dataset & field mode & aggregation & mAP $\uparrow$ & top-1 $\uparrow$ & sec./shape $\downarrow$ \\
\midrule
faust\_reg & heat\_method\_geodesic & pooled & \second{0.122} & \second{0.090} & 0.39 \\
faust\_reg & heat\_method\_geodesic & vlad & \best{0.125} & \best{0.110} & 0.39 \\
tosca & heat\_method\_geodesic & pooled & \second{0.505} & \second{0.613} & 2.00 \\
tosca & heat\_method\_geodesic & vlad & \best{0.555} & \best{0.812} & 2.00 \\
\bottomrule
\end{tabular}
\end{table}

\begin{table}[!htbp]
\centering
\small
\caption{Raw implicit heat-response diagnostics before positive-part clipping and log normalization.  These columns are descriptive rather than accuracy metrics.  The large negative fractions show that cotangent point-source responses are sign-indefinite in this discretization.}
\label{tab:heat-response-diagnostics}
\begin{tabular}{lccccc}
\toprule
dataset & steps & mean neg. frac. & max neg. frac. & min response & max response \\
\midrule
faust\_reg & 1 & 0.473 & 0.541 & -5.08e+04 & 1.66e+04 \\
tosca & 1 & 0.468 & 0.544 & -3.48e+04 & 3.42e+04 \\
\bottomrule
\end{tabular}
\end{table}

\begin{table}[!htbp]
\centering
\small
\caption{One-step versus multistep implicit heat responses inside the DGM pipeline.  mAP and top-1 are higher-is-better ($\uparrow$); runtime is lower-is-better ($\downarrow$).  Best and second-best mAP are marked within each dataset/aggregation block.}
\label{tab:heat-steps}
\begin{tabular}{llcccc}
\toprule
dataset & aggregation & steps & mAP $\uparrow$ & top-1 $\uparrow$ & sec./shape $\downarrow$ \\
\midrule
faust\_reg & pooled & 1 & \best{0.245} & 0.380 & 0.60 \\
faust\_reg & pooled & 2 & \second{0.233} & 0.280 & 0.69 \\
faust\_reg & pooled & 4 & 0.192 & 0.200 & 0.91 \\
faust\_reg & pooled & 8 & 0.148 & 0.110 & 1.38 \\
faust\_reg & vlad & 1 & \best{0.360} & 0.530 & 0.60 \\
faust\_reg & vlad & 2 & \second{0.348} & 0.530 & 0.69 \\
faust\_reg & vlad & 4 & 0.337 & 0.490 & 0.91 \\
faust\_reg & vlad & 8 & 0.272 & 0.420 & 1.38 \\
tosca & pooled & 1 & \second{0.376} & 0.562 & 3.44 \\
tosca & pooled & 2 & \best{0.464} & 0.650 & 4.08 \\
tosca & pooled & 4 & 0.310 & 0.362 & 5.65 \\
tosca & pooled & 8 & 0.195 & 0.250 & 8.71 \\
tosca & vlad & 1 & \second{0.533} & 0.750 & 3.44 \\
tosca & vlad & 2 & \best{0.578} & 0.838 & 4.08 \\
tosca & vlad & 4 & 0.452 & 0.675 & 5.65 \\
tosca & vlad & 8 & 0.278 & 0.375 & 8.71 \\
\bottomrule
\end{tabular}
\end{table}

\begin{table}[!htbp]
\centering
\small
\caption{Moment-set and tensor-branch ablation.  Retrieval metrics are higher-is-better ($\uparrow$). Tensor delta is pooled DGM mAP with tensor invariants minus pooled DGM mAP without them; larger deltas indicate a more helpful tensor branch.}
\label{tab:moment-tensor}
\begin{tabular}{llccc}
\toprule
dataset & moments & VLAD mAP $\uparrow$ & pooled mAP $\uparrow$ & tensor delta $\uparrow$ \\
\midrule
faust\_reg & mean+var & \second{0.324} & \second{0.224} & -0.017 \\
faust\_reg & +skew+kurt. & 0.307 & 0.215 & \best{-0.006} \\
faust\_reg & +min/max & \best{0.344} & \best{0.239} & \second{-0.009} \\
tosca & mean+var & \second{0.490} & 0.349 & -0.006 \\
tosca & +skew+kurt. & 0.439 & \second{0.367} & \best{-0.003} \\
tosca & +min/max & \best{0.575} & \best{0.393} & \second{-0.005} \\
\bottomrule
\end{tabular}
\end{table}

\FloatBarrier
\subsection{Information, Topology, and Seed Diagnostics}
This block collects diagnostics that describe what the seed-conditioned fields retain after moment compression, how the multiscale fields behave as simple filtrations, and whether deterministic seed rules are stable under vertex reordering.

\begin{table}[!htbp]
\centering
\small
\caption{Moment-compression and soft-Voronoi diagnostics for DGM fields.  PCA-6 variance and moment $R^2$ are higher-is-better ($\uparrow$) as compression diagnostics; entropy and margin are descriptive soft-Voronoi statistics rather than accuracy metrics.}
\label{tab:information-compression}
\begin{tabular}{lccccc}
\toprule
dataset & scale & PCA-6 var. $\uparrow$ & moment $R^2$ $\uparrow$ & soft entropy & top1--top2 margin \\
\midrule
faust\_reg & 0.01 & 0.843 & 0.727 & 0.355 & 0.399 \\
faust\_reg & 0.03 & 0.805 & 0.701 & 0.405 & 0.206 \\
faust\_reg & 0.07 & 0.816 & 0.688 & 0.594 & 0.076 \\
faust\_reg & 0.15 & 0.814 & 0.686 & 0.666 & 0.024 \\
tosca & 0.01 & 0.829 & 0.713 & 0.402 & 0.353 \\
tosca & 0.03 & 0.825 & 0.701 & 0.463 & 0.182 \\
tosca & 0.07 & 0.798 & 0.684 & 0.575 & 0.069 \\
tosca & 0.15 & 0.815 & 0.577 & 0.685 & 0.037 \\
\bottomrule
\end{tabular}
\end{table}

\begin{table}[!htbp]
\centering
\small
\caption{0D persistent-homology diagnostic on DGM seed-field filtrations.  Values are mean total persistence over eight shapes per dataset; SMAL-toy is a local synthetic-animal sanity-check set.  The drop from short to long diffusion scales is consistent with the view that DGM scales move from local textured fields to smoother global context.}
\label{tab:ph-diagnostic}
\begin{tabular}{lcccc}
\toprule
dataset & mean $t=.01$ & mean $t=.15$ & var $t=.01$ & var $t=.15$ \\
\midrule
faust\_reg & 1.988 & 0.355 & 1.103 & 0.313 \\
smal\_toy & 1.780 & 0.459 & 0.811 & 0.191 \\
tosca & 3.064 & 0.402 & 2.415 & 0.435 \\
\bottomrule
\end{tabular}
\end{table}

\begin{table}[!htbp]
\centering
\small
\caption{Vertex-reordering stability. Global cosine is higher-is-better ($\uparrow$), while local relative $L_2$ error is lower-is-better ($\downarrow$). Deterministic seed modes are invariant up to numerical precision after aligning local descriptors back to the original vertex order; random Euclidean FPS is not.}
\label{tab:seed-permutation}
\begin{tabular}{llcc}
\toprule
dataset & seed mode & global cosine $\uparrow$ & local rel. $L_2$ $\downarrow$ \\
\midrule
faust\_reg & euclidean\_deterministic & \best{1.000000} & \best{2.57e-12} \\
faust\_reg & euclidean\_random & 0.994013 & 5.60e-01 \\
faust\_reg & geodesic\_deterministic & \best{1.000000} & \second{2.89e-12} \\
tosca & euclidean\_deterministic & \best{1.000000} & \second{5.90e-11} \\
tosca & euclidean\_random & 0.976712 & 7.81e-01 \\
tosca & geodesic\_deterministic & \best{1.000000} & \best{1.52e-11} \\
\bottomrule
\end{tabular}
\end{table}

\FloatBarrier
\subsection{Design Probe and Negative-Insight Summaries}
These tables collect the mechanism probes and negative-evidence lessons that are discussed in the main text.  They are placed here because they summarize secondary protocol layers and design hypotheses; the primary experimental evidence appears in the main-text tables and figures.

\begin{table}[!htbp]
\centering
\small
\caption{Mechanism probes extracted from non-leading experiments.  The probes turn ranking shortfalls into testable statements about descriptor design, solver compatibility, and discretization.}
\label{tab:mechanism-probes}
\begin{tabular}{p{0.20\linewidth}p{0.27\linewidth}p{0.41\linewidth}}
\toprule
Probe & Result & Interpretation \\
\midrule
Noisy heat-method fields &
FAUST heat-method VLAD improves from $0.125$ to $0.133$ mAP at $\sigma=0.01$; TOSCA VLAD improves from $0.534$ to $0.555$ at $\sigma=0.20$. &
Small perturbations can improve precise distance fields, but the gain is much smaller than the gap to implicit DGM.  Useful roughness is therefore structured, not arbitrary noise. \\
Cross-shape alignment &
Linear CSAS $R^2$ stays high for WKS, HKS, GMSD-WKS, and SI-HKS, is lower for GMSD-HKS, and is lowest for DGM. &
DGM can be locally discriminative while violating the coordinate-comparability assumption used by functional maps; CSAS diagnoses this interface independently of retrieval accuracy. \\
Symmetry seed curve &
Euclidean seed frames reach $0.868$--$0.887$ side balanced accuracy at $k=4$ and decrease mildly with larger $k$; geodesic seeds rise from $0.650$ at $k=4$ to about $0.81$ by $k=16$. &
Symmetry information is a controllable design axis governed by the seed frame, not simply by the number of seeds. \\
Negative heat values &
Raw implicit responses are negative for $47.5\%$ of FAUST samples and $46.2\%$ of TOSCA samples; correlations with valence, edge-length variation, aspect ratio, and area are near zero. &
The sign pattern is not just a local bad-triangle artifact.  The practical ``heat'' response is a signed cotangent-resolvent signal. \\
Decimated FAUST &
After $50\%$ simplification with random geometric jitter, implicit DGM-VLAD remains $0.344$ mAP while heat-method VLAD is $0.136$; the simplified meshes have 100 distinct topology hashes. &
The implicit-field advantage is not explained solely by the original shared FAUST tessellation, although benchmark homogeneity still affects the magnitude of the effect. \\
\bottomrule
\end{tabular}
\end{table}

\begin{table}[!htbp]
\centering
\small
\caption{Negative results as design evidence.  Each row records a plausible shortcut that the audit tested and the design lesson suggested by the observed failure mode.}
\label{tab:negative-insights}
\begin{tabular}{p{0.25\linewidth}p{0.31\linewidth}p{0.32\linewidth}}
\toprule
Question tested & Observation & Design lesson \\
\midrule
Can cheaper heat-kernel approximations replace eigensolvers directly? &
They are faster in this implementation but remain far below WKS and GMSD-HKS in matched retrieval. &
Runtime alone is not the right proxy; the descriptor signal induced by the approximation must be audited downstream. \\
Does a more faithful or smoother heat evolution always help DGM? &
Multistep diffusion is non-monotonic: one step is best on FAUST-Reg, while TOSCA improves briefly and then degrades. &
DGM benefits from finite-time contrast, not from the smoothest heat field available. \\
Can a simple learned linear substitute beat hand normalization? &
Linear whitening improves the VLAD variant but not the pooled variant. &
Normalization and aggregation interact; they should be evaluated as a coupled block. \\
Do lifted foundation features dominate classical geometry descriptors? &
Simple DINOv2 lifting remains below WKS, DGM, and GMSD-HKS on full FAUST-Reg retrieval. &
Pretrained image features need geometry-aware lifting and aggregation before replacing surface descriptors. \\
Does nearest-neighbor descriptor quality imply functional-map quality? &
DGM gives the best nearest-neighbor correspondence but the worst result after functional-map refinement. &
Descriptor coordinates must be aligned with the assumptions of the solver, not only locally discriminative. \\
\bottomrule
\end{tabular}
\end{table}

\FloatBarrier

\subsection{Symmetry and Foundation-Feature Diagnostics}
The final block reports two bounded probes that are useful for interpretation but not central enough to lead the paper: whether DGM carries side information on symmetric shapes, and whether a simple lifted image-foundation feature baseline replaces geometry-specific descriptors in this protocol.

\begin{table}[!htbp]
\centering
\small
\caption{Left/right side proxy diagnostic on 20 FAUST-Reg meshes.  Balanced accuracy and receiver-operating-characteristic area under the curve (ROC-AUC) are higher-is-better ($\uparrow$).  This is not a correspondence benchmark; it measures whether a descriptor carries symmetry-informative side information.}
\label{tab:symmetry-side}
\begin{tabular}{lccc}
\toprule
method & shapes & balanced acc. $\uparrow$ & ROC-AUC $\uparrow$ \\
\midrule
DGM & 20 & \best{0.836 $\pm$ 0.057} & \best{0.906 $\pm$ 0.049} \\
WKS & 20 & \second{0.547 $\pm$ 0.064} & \second{0.562 $\pm$ 0.071} \\
GMSD-HKS & 20 & \second{0.547 $\pm$ 0.058} & 0.561 $\pm$ 0.072 \\
HKS & 20 & 0.526 $\pm$ 0.039 & 0.533 $\pm$ 0.060 \\
\bottomrule
\end{tabular}
\end{table}

\begin{table}[!htbp]
\centering
\small
\caption{Foundation-feature probe on the same 100 FAUST-Reg shapes.  Retrieval metrics are higher-is-better ($\uparrow$).  Lifted DINOv2 features improve over rendered normal features, but geometry-specific descriptors remain stronger in this zero-shot lifting protocol.}
\label{tab:foundation-probe}
\begin{tabular}{lcccc}
\toprule
method & shapes & dim & mAP $\uparrow$ & top-1 $\uparrow$ \\
\midrule
WKS & 100 & 72 & \best{0.290} & \best{0.370} \\
DGM & 100 & 144 & \second{0.237} & \second{0.310} \\
GMSD-HKS & 100 & 18 & 0.222 & \second{0.310} \\
DINOv2-small lift & 100 & 1152 & 0.151 & 0.200 \\
normal lift & 100 & 9 & 0.119 & 0.090 \\
\bottomrule
\end{tabular}
\end{table}

\FloatBarrier

\end{document}